\documentclass[lettersize,journal]{IEEEtran}
\usepackage{amsmath,amsfonts}
\usepackage{algorithmic}
\usepackage{algorithm}
\usepackage{array}
\usepackage[caption=false,font=normalsize,labelfont=sf,textfont=sf]{subfig}
\usepackage{textcomp}
\usepackage{stfloats}
\usepackage{url}
\usepackage{verbatim}
\usepackage{graphicx}
\usepackage{cite}
\usepackage{color}
\usepackage{booktabs} 
\usepackage{multirow}
\usepackage{graphicx}
\usepackage{subcaption}
\usepackage{hyperref} 
\definecolor{orange}{RGB}{255, 128, 0}

\begin{document}

\title{EA-RAS: Towards Efficient and Accurate End-to-End Reconstruction of Anatomical Skeleton}

\author{Zhiheng Peng, Kai Zhao, Xiaoran Chen, Li Ma, Siyu Xia, Changjie Fan, Weijian Shang, Wei Jing
        % <-this % stops a space
\thanks{Corresponding author: Wei Jing. Email: jingwei04@corp.netease.com}
\thanks{Zhiheng Peng, Kai Zhao, Xiaoran Chen, Li Ma, Changjie Fan, Weijian Shang and Wei Jing are with NetEase Fuxi Robot Department, HangZhou, China. (E-mail: pengzhiheng1202@163.com; zhaokai02@corp.netease.com)}
\thanks{Siyu Xia is with School of Automation, Southeast University, China.}

}



\maketitle

\begin{abstract}
Efficient, accurate and low-cost estimation of human skeletal information is crucial for a range of applications such as biology education and human-computer interaction. 
However, current simple skeleton models, which are typically based on 2D-3D joint points, fall short in terms of anatomical fidelity, restricting their utility in fields.
On the other hand, more complex models while anatomically precise, are hindered by sophisticate multi-stage processing and the need for extra data like skin meshes, making them unsuitable for real-time applications. To this end, we propose the EA-RAS (Towards Efficient and Accurate End-to-End Reconstruction of Anatomical Skeleton), a single-stage, lightweight, and plug-and-play anatomical skeleton estimator that can provide real-time, accurate anatomically realistic skeletons with arbitrary pose using only a single RGB image input. Additionally, EA-RAS estimates the conventional human-mesh model explicitly, which not only enhances the functionality but also leverages the \emph{outside} skin information by integrating features into the \emph{inside} skeleton modeling process.
In this work, we also develop a progressive training strategy and integrated it with an enhanced optimization process, enabling the network to obtain initial weights using only a small skin dataset and achieve self-supervision in skeleton reconstruction.
Besides, we also provide an optional lightweight post-processing optimization strategy to further improve accuracy for scenarios that prioritize precision over real-time processing.  The experiments demonstrated that our regression method is over 800 times faster than existing methods, meeting real-time requirements. Additionally, the post-processing optimization strategy provided can enhance reconstruction accuracy by over 50\% and achieve a speed increase of more than 7 times.

\end{abstract}

% \begin{IEEEkeywords}
% Article submission, IEEE, IEEEtran, journal, \LaTeX, paper, template, typesetting.
% \end{IEEEkeywords}

\section{Introduction}
\IEEEPARstart{E}{stimating} and modeling the human skeletal system is crucial for various applications\cite{balmik2022nao,liu2022spatial,ijcai2021-0135}, including robotics, gaming, person re-identification, and etc. 
Most \textbf{skeletal reconstruction} work uses interconnected keypoints with lines to represents human skeleton~\cite{cao2017realtime,xu2022vitpose,sun2019deep}. 
This representation enables the recognition of semantic postures and actions such as sitting, standing, and running. However, the simple representation fails to describe the anatomical details of real human skeletons. For applications in biological fields like physical therapy robots\cite{hu2021novel,xu2024toward}, biological education\cite{stevens2024bioclip}, and game simulations\cite{siyao2022bailando} that requires accurate skeletal reconstruction, the abovementioned methods could be insufficient. 
As is well known, obtaining the accurate skeleton information requires the medical equipment, such as dual-energy X-ray absorptiometry (DXA). But such an equipment is expensive, poses radiation risks, and is not suitable for use in daily environments. Therefore, researchers have introduced \textbf{anatomical skeletal reconstruction}. Various methods can be used to reconstruct the anatomical structure through modeling and simulation~\cite{saito2015computational,ali2013anatomy,Keller:CVPR:2022,kadlevcek2016reconstructing}. 
While incorporating complex anatomical constraints to ensure realism, these methods have encountered a trade-off, leading to an inability to fulfill real-time processing requirements.
To address these limitations, this work focus on \textbf{lightweight anatomical skeletal reconstruction}, which uses low-cost visual methods to generate accurate skeletal reconstruction results at a faster speed. 

\begin{figure}[t]
\centering
\includegraphics[width=\linewidth]{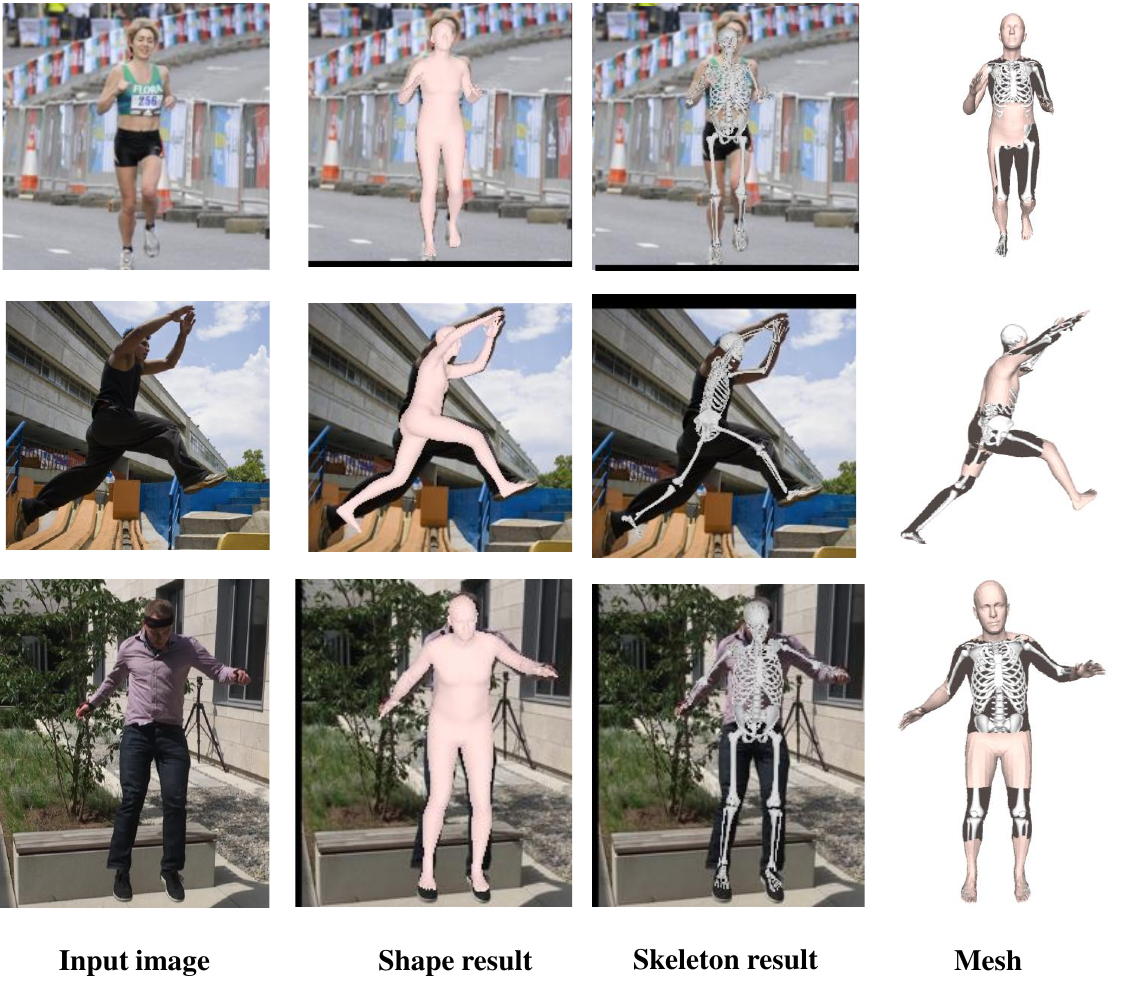}
\caption{Our method completes both human skin and anatomical skeleton reconstructions. The model achieves good inference results on 3DPW \cite{von2018recovering}and the lower resolution LSP \cite{johnson2010clustered} dataset.}
\label{introduc}
\end{figure}

A primary challenge in employing conventional end-to-end training methods for lightweight anatomical skeletal reconstruction is the collection of data. Due to privacy concerns and medical ethical restrictions, there is currently a lack of large-scale real data mapping DXA images to anatomical skeletons.
Moreover, because of the complex structure of the actual skeleton and the connection between bones and skin, manually labeling RGB images directly can be challenging.

In response to the abovementioned limitations, several \emph{multi-stage} methods can also successfully achieve this task without using the end-to-end \emph{single-stage} method.
For instance, researchers can predict human skin mesh from images using methods like \cite{zimmermann2017learning,xu2019denserac,guler2018densepose,loper2014mosh,mahmood2019amass,lassner2017unite}. After obtaining the mesh, the intrinsic skeleton can be optimized using anatomical geometric information\cite{Keller:CVPR:2022}.
However, these multi-stage methods divide the task into two stages: first from RGB image to skin, then from skin to bones. 
Although this design simplifies the difficulty of directly reconstructing skeletons from RGB images, multi-stage methods with separation of modules bares the risk of information loss across stages, error accumulation and feature misalignment due to the isolation between bones and original images.

To address the abovementioned challenges, this work aims to design a \emph{single-stage}, lightweight, plug-and-play method that accurately estimates the anatomical skeleton of the human body in real-time, with a single RGB image input. 
Compared to these conventional simplistic bone representations\cite{cao2017realtime,xu2022vitpose,sun2019deep}, the bone model we employed is more realistic and takes into account anatomical constraints, leading to a better representation of human kinematics. To overcome data scarcity,  inspired by \cite{kolotouros2019learning},  we combine the advantages of regression and optimization, and developed a semi-supervised progressive training method. Compared to the direct data-driven approach, this method enables the network to train the weights using only a mini-batch skin dataset and then achieve stable and rapid convergence with better generalization. 
In addition, we also provide a lightweight post-processing optimization method for higher accuracy needs. Moreover, we employ a dual-branching approach to better consider and utilize the similarities and differences between bones and skin by sharing information and imposing mutual constraints.
Qualitative results are shown in Figure \ref{introduc}, and the project is available at \href{https://ea-ras.github.io}{https://ea-ras.github.io}.

The contributions of this work can be summarized as follows:
\begin{itemize}
\item We introduce EA-RAS, a lightweight single-stage estimation method, inferring the skeleton of a person in arbitrary pose from a single RGB image. We also design an optional post-processing optimization method to further improve accuracy for scenarios that prioritize precision over real-time processing. To the best of our knowledge, we are the first to predict both the inside (bones) and the outside (skin) with one single-stage model. 

\item We propose a dual-branch mechanism to fuse skin information in the skeleton estimation process, avoiding the information loss and inaccuracies that exists in multi-stage methods.

\item We propose a three-stage progressive training method to eliminates the need for complex full skeleton annotation and ensure fast convergence. Only relies on a small skin dataset\cite{johnson2010clustered} to train the initial weights in the first stage, our training method can achieve self-supervision subsequently through a enhanced optimization process called OSF$^{+}$, which is tailored specifically for training purposes. Experiments have shown high accuracy and speed, along with good performance on a large-scale dataset\cite{von2018recovering} that has not been previously utilized.
\end{itemize}
\section{Related Work}
\subsection{Skeleton Reconstruction by Keypoints}
Existing skeletal reconstruction methods mainly emphasize action recognition and pose estimation, rather than predicting the physiological state of bones. The typical method for representing poses is to display human joint positions through 2D keypoints that show body movement.
Earlier exploration of human pose heavily relied on handcrafted features~\cite{andriluka2009pictorial,gkioxari2013articulated}. Recently, many deep learning-based methods have been employed to significantly enhance the performance of human pose estimation~\cite{yi2022human,zou2021modulated,zhang2021adafuse}. In terms of network architecture, these methods are mainly divided into two categories: single-stage methods~\cite{papandreou2017towards,he2017mask} and multi-stage methods~\cite{newell2017associative,wang2018stacked}. Single-stage methods primarily involve training a backbone network for image classification to generate heatmaps or corresponding offsets for human keypoints, thereby obtaining the final estimation of keypoints~\cite{he2017mask}.  Multi-stage methods predict image keypoints and then associating these keypoints with human instances~\cite{carreira2016human}. Wei \emph{et al.} \cite{wang2018stacked} use detectors to locate individuals, followed by using convolutional encoders to accomplish joint dependency and sequential prediction\cite{newell2017associative,wang2018stacked}.
As the emerging of transformers, the above method has also %achieved exciting performance improvements
achieved significant performance improvements.
 TransPose\cite{yang2021transpose} utilizes CNN-extracted features to directly capture the overall relationship. TokenPose\cite{li2021tokenpose} introduces additional tokens to estimate occluded keypoint locations and model inter-keypoint relationships through token-based representations. However, these methods require a large amount of annotated data for training.
Although current 2D keypoints exhibit good performance, 
%the mere positioning of a few keypoints makes it challenging to articulate the internal skeletal structure and decompose human posture effectively.
the mere keypoints makes it difficult to construct the internal skeleton structure effectively.

\subsection{Skeleton Reconstruction by Anatomical Models}

Most approaches to obtaining skeletal anatomical models are achieved by supervising the sparse correspondence between the outer skin and the inner part of the skeleton, such as \cite{besl1992method,bronstein2008numerical}. Some work achieves a dense displacement field for the physical pose by completing regularization \cite{mateus2008articulated,lipman2009mobius}.
AT~\cite{ali2013anatomy} proposes the first semi-automated method for creating anatomy, passing it into the target skin while preserving the skeletal structure, and mapping the internal anatomy of the source model by harmonic deformation. Based on this, \cite{bauer2016modelisation} adds constraints to position the bones inside.
Although these methods have shown 
the more detailed approach for local modeling, the complex constraints, displacement fields and the complex calculation of interpolation processing have brought serious shortcomings in universality. %In addition, there is no guarantee of the reality of the obtained bones.
In addition, these methods are not validated on real data. 
OSSO~\cite{Keller:CVPR:2022} uses the DXA scan data to convert the body shape model to the inside skeleton. However, the body model must be obtained in advance, and the pose adaption requires a large optimization time.

Although anatomically informed skeletal reconstruction techniques offer high fidelity in depicting physiological skeletal structures, the high computational demands and prerequisite body model specifications they require make their application in real-time settings challenging.

\subsection{Human Body Recovery}
%As mentioned above, 
As mentioned in previous section, inferring the anatomical skeleton may require human body models and prior information. There are some methods to recovery the human body from RGB images, roughly divided into two categories: optimization-based and regression-based.
To estimate a 3D body skin mesh that contains human pose and shape, the optimization-based approach involves forecasting the pertinent parameters and necessitates alignment with two-dimensional observational data.

Early work~\cite{balan2008naked,hasler2010multilinear} based on SCAPE restricted the area of pixels and punished non-overlapping conditions. SMPLify~\cite{bogo2016keep} minimizes the reprojection error between 2D joints and SMPL joints and uses several regularization terms to keep the joint rotation and shape naturally. However, the optimization-based method is sensitive to the initial value and has iteration time overhead. Xiang \emph{et al.} \cite{8953618}, Hassan \emph{et al.}~\cite{9010321}, Zhang \emph{et al.} \cite{10.1007/978-3-030-58610-2_3} analyzed the relationship between the image and the objects related to the human body. The regression-based method directly predicts the human body model through deep learning. The parametric output methods regress the parameters of models and use them to reconstruct the human body~\cite{kocabas2021pare}. The non-parametric methods predict the mesh vertices~\cite{choi2020pose2mesh,hanocka2020point2mesh} or body shape~\cite{varol2018bodynet}, and then fit them to a human model~\cite{wang2018pixel2mesh}. With the development of Transformer\cite{vaswani2017attention} in recent years, many human reconstruction methods based on it have also achieved good results \cite{vaswani2017attention, lin2021end-to-end, lin2021-mesh-graphormer}.
Moreover, some work has implemented more effective supervision and introduced alignment constraints, such as meshes aligned~\cite{zhang2021pymaf}, surface landmarks~\cite{lassner2017unite}, pose keypoints and contours~\cite{Pavlakos2018}, semantic part segmentation~\cite{Omran2018} or raw pixels~\cite{kanazawa2018end}.

Although these methods provide %valuable
significant skin information, errors in the body recovery process can interfere with the subsequent skeleton reconstruction, leading to inaccuracies. Additionally, the statistical relationship between the skin and the internal skeleton has not been fully taken into account in the body recovery stage, potentially resulting in information loss during the skeleton reconstruction stage.

\section{Methodology}
\begin{figure*}[t]
\centering
\includegraphics[width=\linewidth]{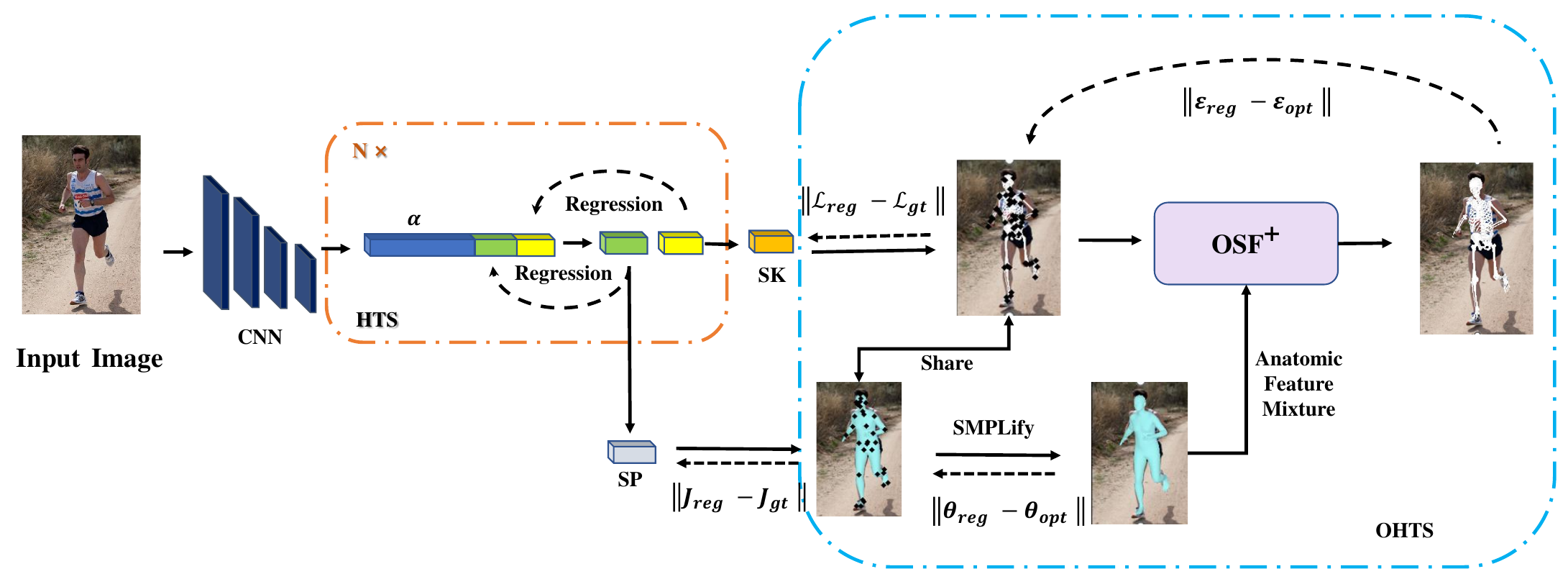}
\caption{An overview of the entire proposed method. Our method employs an iterative optimization approach to achieve self-supervision of regression results. The regression network is augmented with our designed HTS feature, and the network predicts the parameters of human SMPL \cite{SMPL:2015} and skeleton model \cite{Keller:CVPR:2022}. Skeleton and human body features are integrated and complemented through the OHTS in the process of optimization supervision.}
\label{baseline}
\end{figure*}

As shown in Figure~\ref{baseline}, the proposed one-stage method uses optimization to achieve self-supervision of regression results and efficiently complete the parameters prediction of human and skeleton. The human and skeleton information is fused both in regression (HTS) and optimisation (OHTS).

\subsection{Human Modeling}
\textbf{Human Body Model:}
SMPL \cite{SMPL:2015} provides a parameterized human model $HM(\beta_{body},p )$ with a learned human skin.  The reconstruction is completed with 6,890 vertices using the pose parameter $p$ and the shape parameter $\beta_{body}$. 

\begin{figure}[t]
\centering
\includegraphics[width=0.9\linewidth]{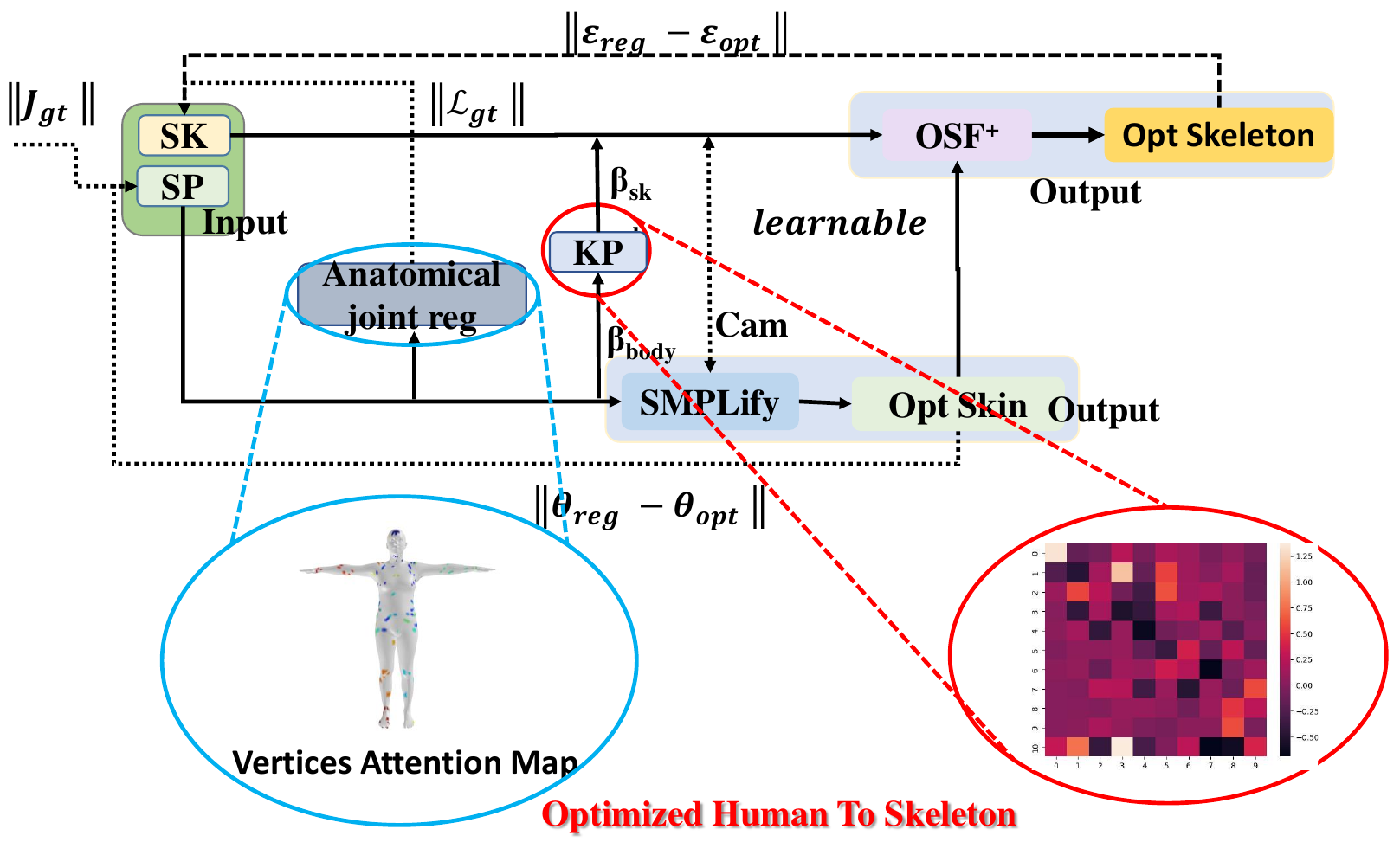}
\caption{An overview of the OHTS. Keypoints serve as implicit supervision in the regression process. The results of the optimization phase provide explicit supervision of the regression results. Anatomical joint reg and KP helps to get the keypoints and shape of skeleton. Colored regions represent vertices affecting skeletal keypoints, while the heatmap indicates the correlation between body shape and skeletal parameterization.}
\label{OHTS}
\end{figure}

\textbf{Anatomical Skeletal Model:}
In order to move, rotate and scale the bones freely and flexibly, we use the stitched puppet \cite{sigal2007combined} model with anatomical skeleton. The stitched puppet provides an ideal graphic model formula and 
is conducive to optimization under anatomical constraints. Skeleton model $SM(\beta_{skel},t,r)$ can be obtained by providing the skeleton shape parameter $\beta_{skel}$, the translation parameter $t$ and rotating parameter $r$. Similar to the human body model, the skeleton vertexes can determine the position of skeleton joints.

\subsection{Network Structure}

\subsubsection{\textbf{Human To Skeleton (HTS) Matching Module}}

\paragraph{Module Structure} 
To make the model understand the mutual inference relationship between the anatomical skeleton and the human body, we designed the HTS (human body to skeleton) module.

As shown in Fig~\ref{baseline}, the Convolutional Neural Network (CNN) is utilized to extract features from RGB images, generating shared features ($\alpha$). Subsequently, $\alpha$ is combined with parameter information (skeleton and human). 
Initially, these features are initialized from the mean parameter model. The combined feature is then forwarded to a regressor to obtain the human and skeleton parameters. These parameters are also fed back to the combined feature for further iterations. This iterative process can be repeated N times to derive the ultimate human ($SK$) and skeleton parameters ($SP$).
In more detail, the camera parameters $cam$ are the same for both the skin and bones. These parameters are integrated into the human body parameters, denoted as $SP = \{cam, \beta_{body}, p\}$, and $SK = \{t, r\}$. The parameter ($\beta_{skel}$) for adjusting the shape of the skeleton ($SM(\beta_{skel}, SP)$) is generated by $\beta_{body},$ with further details described subsequently.

The parameter extraction of human body is detailed as:

\begin{align}
\theta ^{i}_{reg} = 
    R_{body}\left(concat\left(\alpha,\theta^{i-1} _{reg}\right)\right)
\end{align}
where $R_{body}$ is the self-regressor for the human body branch in HTS. $\theta^{0}_{reg}$ represents the initial parameter value of standard human and initial camera parameter. ${i}\in{[1,3]}$ means current number of regression iterations. 

Similarly, deviation from the human body parameters is to obtain the skeleton parameter:
\begin{align}
\varepsilon^{i} _{reg} = 
    R_{skeleton}\left(concat\left(\alpha,\theta^{i-1} _{reg},\varepsilon^{i-1} _{reg}\right)\right)
\label{HTS-skel-formula}
\end{align}
where $R_{skeleton}$ is the self-regressor for skeleton parameter in HTS, and $\varepsilon_{reg}$ is the skeleton parameter in $SK(.)$ including the translation $t$ and rotation $r$ parameters. $\varepsilon^{0}_{reg}$ is set to zero before the training. 

\paragraph{Bidirectional Adjustment}

The human body parameters are obtained through the regression network, and the loss function is:
\begin{equation}
L_{hum} = \lambda _{hj} \left \| J_{reg} - J_{gt} \right \| + L_{\theta}
\label{human_body_regression_format}
\end{equation}
where $\lambda_{hj}$ is the weight. $J_{reg}$ represents the 2D joints obtained by reprojecting the 3D joints to the image using the predicted camera parameters. 
The $J_{reg}$ is forced to align with $J_{gt}$ which represents the ground truth 2D joints. $L_{\theta}$ is the straight supervision of human parameters calculated in the following optimization step.

In addition, we can also perform the regression to infer the skeleton joints position to provide posture supervision. However, rather than the common human keypoints in human body model, the anatomical skeleton joints are more complicated. So it is difficult to control the skeleton by simple constraints. For this reason, we limit the results of the components and add the weak supervision process of Mean Square Error (MSE):

\begin{align}
\label{skeleton_regression_format}
\begin{split}
L_{skel} &= \lambda _{l} L(\mathcal{L}_{reg}, \mathcal{L}_{gt})  + \lambda _{r}L(r_{reg}, r_{gt}) 
\\&+ \lambda _{t}L(t_{reg}, t_{gt})
\end{split}
\end{align}

\noindent where $\mathcal{L}_{reg}$ and $\mathcal{L}_{gt}$ are the predicted 3D joints and groundtruth 3D joints which can be inferred directly by the human body model, respectively. $L(t_{reg}, t_{gt})$ and $L(r_{reg}, r_{gt})$ are weak supervision of bone translation and rotation.  $L(t_{reg}, t_{gt})$ is smooth L1 loss and $L(r_{reg}, r_{gt})$ is MSE loss. 

The following optimization step generates ground truth. Compared with learning the human body model, skeleton need more parameters to supervised. Multi-supervised parameter is very likely to lead the network into local minimal, causing local minimum and instability during the learning process. In order to make the supervision more effective, we designed progressive training. During training. The weights of $\lambda_{r}$ and $\lambda_{t}$ are gradually adjusted, allowing the model to learn more deeply about statistical correlations between the skeleton and the human body. Training details are explained in the following sections.

In order to ensure the effective learning of skeletal features within the spatial feature retrieval domain, straight guidance is imposed. We separate the output results and incorporate distinct supervisory signals for both skeletal and human body information, facilitating the joint learning within the network of mutual information. This enable our model to simultaneously acquire relevant content for human body reconstruction.

\subsubsection{\textbf{Optimized Human To Skeleton (OHTS)}}

\paragraph{Module Structure}

As shown in Figure~\ref{OHTS}, the human body ($SP = \theta_{reg}$) and skeletal parameter ($SK = \varepsilon_{reg}$) obtained by HTS is recombined and supervised in this process.

In the figure, solid and dashed lines represent the forward chaining and the supervision processes, respectively. To be more specific, we construct the human body model $HM(\theta_{reg})$ and skeleton model $SM(\varepsilon_{reg})$ through the HTS module. Then the $HM(.)$ generates body keypoints $J$ and projects them to the image using camera parameter $cam$. The 2D ground truth $J_{gt}$ of the skin dataset provides supervision. Meanwhile, skeleton keypoints $\mathcal{L}$ are supervised by the ground truth $\mathcal{L}_{gt}$ by \textbf{Anatomical joint reg} module.

This module, derived from work\cite{Keller:CVPR:2022}, trained with real data, and map human body $HM(.)$ to keypoints $\mathcal{L}$ through Vertices Attention Map.
Similarly, the shape parameters ($\beta_{skel}$) of the skeleton ($SK$) is 
inferred through the keypoint prediction matrix (\textbf{KP}): 
\begin{align}
\beta_{skel}^{reg}= KP(\beta_{body})
\end{align}
It is learned and updated continuously throughout the process. 
Figure~\ref{OHTS} visualizes the Vertices Attention Map and the heatmap of KP. Meanwhile, camera parameters ($cam$) is shared between the human body and skeleton. The parameters of skin and bones are optimized using the following methods:

\begin{itemize}
\item SMPLify: Using $\theta_{reg}$ as initial value, optimize $\theta_{reg}$ to $\theta_{opt}$ and obtain the \textbf{Opt Skin} by $HM(.)$.
\item OSF$^{+}$: Using $\varepsilon_{reg}$ and $\beta_{skel}$ as initial value, obtain the \textbf{Opt skeleton} by Opt Skin, and optimize $\varepsilon_{reg}$ to $\varepsilon_{opt}$.
\end{itemize}
$\theta_{opt}$ and $\varepsilon_{opt}$ are used to supervise the regression result of HTS. These two methods are detailed in the following self-supervision module.

\paragraph{Skeleton Parameterized}

As shown in the Figure~\ref{midinpaper} (Parameter Decomposition), the entire human skeletal structure $S$ is composed of multiple skeletal blocks $S_n$. We use a vertex skeletal selection regressor on $S$ to obtain Matching Points ($MP$) and Control Points ($CP$)  (orange region), which are specific requirements to satisfy the skeletal motion. Specifically, $MP$ is used to control its connection with neighboring skeletal blocks through clustering. The entire skeleton is constructed by connecting the $MP$ on different bone blocks. $CP$ is used to control the posture of different skeletons. By controlling the bone block coordinate system centered at $CP$, each vertex of the skeletal block is adjusted. The red dashed circle visualizes $MP$ and $CP$ in two bone pieces, as well as the connection relationship between the bones (taking $S_N$ and $S_{N+1}$ as examples). The topological structure of the entire skeletal area is controlled by multiple topological control points, namely keypoints $\mathcal{L}$. 

This parameterized topological skeleton enhances the controllability of anatomical constraints. Each skeletal block has an independent coordinate system. As indicated by the blue cuboid in the figure, an implicit linking relationship is established between the skeletal coordinate systems through $CP$. The rotation ($r$), translation ($t$), and scaling ($\beta$) can control the coordinate system. The entire parametric process and optimization can be computed in parallel and is real-time differentiable. The visualization in the Figure~\ref{midinpaper} shows the effects of different control parameters.

\subsubsection{\textbf{Self-Supervision}}

Our model performs multi-steps joint optimization and supervision processes for the skeleton learning process. In order to achieve self-supervised single-stage methods, we generate regression supervision through optimization methods. The optimization only requires the rule-based information to generate the supervised values for reconstruction. By employing different optimization methods for the human body output and skeleton output, we achieve complementary advantages of both optimization and regression. This enables one-stage inference and training for the skeleton without the need for additional labels.

\textbf{SMPLify}

After the human body regression process, an optimization process is adopted to supervise the related parameter directly. Here we use the SMPLify \cite{bogo2016keep}, which is similar to the work of SPIN \cite{kolotouros2019learning}. 
As in Equation~\ref{human_body_regression_format}, the optimization process is used to obtain the $L_{\theta}$, which can be calculated as follows: 
\begin{align}
L_{\theta } = \left \| \theta _{reg} - \theta _{opt} \right \| 
\end{align}
where $\theta_{opt}$ can obtained by this optimization algorithm.

\textbf{Obtain Skeleton Fast (OSF$^+$)}

\begin{figure*}[t]
\centering
\includegraphics[width=\linewidth]{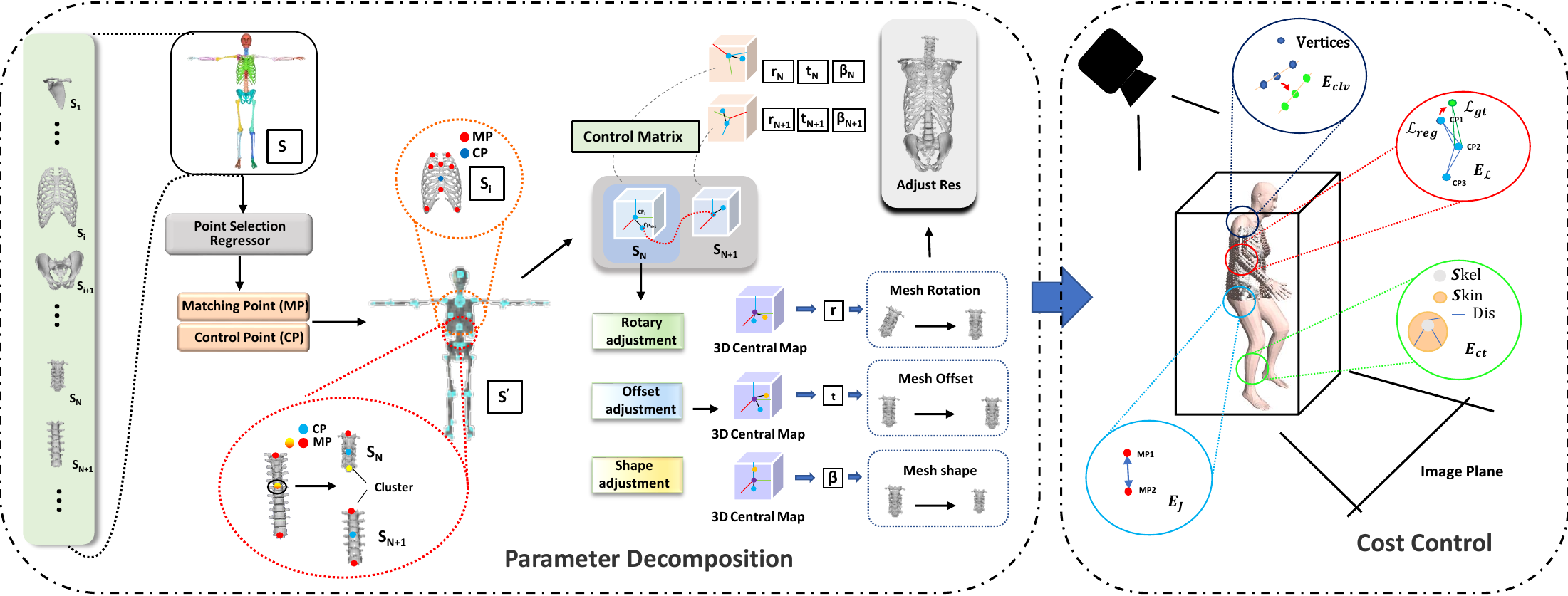}
\caption{\textbf{Parameter Decomposition:} The paramater is optimized simultaneously to control the posture of the skeleton shapes. Match points (MP) and control points (CP) can control joint and position to fit topological cluster. The entire skeleton's topology cluster can be optimized using the corresponding control points.
The cuboid represents the skeletal coordinate system, linking control points to form skeletal morphology.
\textbf{Cost Control} showcases how our optimization process controls the skeleton by maintaining the relationship between vertices and control points (CP). It also ensures that the skeletal and human body mesh points align through cross-sections and maintains the spacing relationship between mesh points (MP) during optimization.}
\label{midinpaper}
\end{figure*}

Inspired by OSSO \cite{Keller:CVPR:2022}, We propose the OSF  (Obtain Skeleton Fast) to speed up the optimization. With the \emph{T-pose} state skeleton as the initial state, the whole process of OSSO passes through three stages: register body point cloud to model, inferring the skeleton to lying human, and jointly optimize body with skeleton to the input pose. 
This approach incurs significant overhead and presents challenges when integrated into regression-based networks.
The proposed OSF uses the skeleton regression prediction as the initial value, which is a single-stage process, and extracts the important anatomical constraints mentioned in OSSO with the transfer puppet loss \cite{zuffi2015stitched}. The final objective function is: 

\begin{align}
\begin{split}
E(t,r) &= \lambda _{l}\left \| \mathcal{L}_{skel}(SM(\beta _{skel}^{opt}, t, r)) - {\mathcal{L}}_{body}\left(HM(\beta_{body},p) \right)  \right \| 
       \\&+ \lambda _{ct}E_{ct}(\beta_{skel}^{opt},t,r; SM_{0})+ \lambda _{j}E_{j}(\beta_{skel}^{opt},t,r;SM_{0}) 
       \\&+ \lambda _{clv}E_{clv}(\beta_{skel}^{opt},t,r;SM_{0})
\end{split}
\label{skeleton_optimize_formula}
\end{align}
where $\mathcal{L}_{skel}(.)$ represents the joints obtained from the skeleton, and ${\mathcal{L}}_{body}(.)$ represents the joints inferred from the human body mesh. $SM_{0}=SM(\beta_{skel}^{opt},t_{0},r_{0})$ represents the skeleton with the initial standard \emph{T-pose} and the inferred shape. $E_{ct}(.)$ is the anatomical constraint which is designed of more than 3,000 points to restrain the relative distance of skeletons \cite{Keller:CVPR:2022}. $E_{j}$ is the penalization term to ensure the same connection distance between different bones. $E_{clv}$ means the anatomical constraints of the clavicle relative to the thorax. $\lambda$ represents corresponding weight coefficients. 
The first term in Equation~\ref{skeleton_optimize_formula} establishes a robust linkage between the skeletal structure and the human form. Subsequent terms concentrate on the skeleton's intrinsic properties, compelling it to adhere to anatomical constraints.
In Figure~\ref{midinpaper} (Cost Control), we visualize the constraint about control point connections  (red), human body cross-sectional view (green), MP (blue), and comprehensive vertices (deep blue), respectively.

In this function, we optimize the shape parameters of skeleton $\beta_{skel}^{opt}$ to represent the corresponding shape relationship between different human body and skeletons, which is achieved through the following formula:
\begin{align}
\beta_{skel}^{opt}=\beta_{skel}^{reg}+\gamma (\beta_{min}+\beta_{max})/2
\end{align}
where $\beta_{min}$ and $\beta_{max}$ are the maximum and minimum parameters for the skeleton model $SM(.)$. $\gamma$ is optimized to adjust vertex displacement between the different shapes. As mentioned above, the KP which to get $\beta_{skel}^{reg}$ is also optimized.

For the exploration of this method, we progressively designed four different optimization processes (OSSO*, OSSO**, OSF, OSF$^{+}$) to get the skeleton from the obtained human body. The details of the four specific methods are introduced in the experiment section. Here we only briefly introduce the OSF by Formula \ref{skeleton_optimize_formula} and OSF$^{+}$ used in our model training and prediction method. OSF used the model parameters obtained by network regression, while OSF$^{+}$ is an extra option to improve the accuracy and speed. It removes supervision of the last three terms in Formula \ref{skeleton_optimize_formula} and optimizes only once after the regression. It should be noted that without a reasonable initial value, OSF$^{+}$ can not obtain accurate results through only once iteration. Fortunately, the regression process described above provide the needed initial optimization value. In this way, the advantages of regression and optimization methods can be effectively combined. Specific analysis and experiments are introduced later.
Through the final optimization results, the needed regression skeleton parameters including $r_{gt}$, and $t_{gt}$ mentioned in Formula \ref{skeleton_regression_format} can be obtatined to supervise the skeleton regressor branch.
For scenarios that prioritize precision over real-time processing, we support a PLUS version based on OSF, conducting multi-round optimization, incorporating the optimization of $\beta_{skel}$, and treating the matrix in $KP(.)$ as one of the optimization variables.

\subsection{Training and Inference}
The entire training process is carried out in an end-to-end method. Specifically, the entire loss function satisfies the following formula.
\begin{align}
\begin{split}
L_{entire} = L_{hum} + L_{skel}
\end{split}
\end{align}
where $L_{hum}$ is the loss of human mesh and $L_{skel}$ is the loss of skeleton mesh.

To improve the train efficiency, we propose a progressive training method, including the position demonstration stage  (stage one), weak supervision stage (stage two), and collaborative improvement stage (stage three). 

In stage one, we train the model using a subset of a smaller dataset, and use the final value as the ground truth after multiple iterations of skeleton optimization. In this stage, model convergence quickly about the skeleton roughly position. During the weak supervision stage, we gradually weaken the last two items in Formula \ref{skeleton_regression_format} to allow the model to learn the statistical correlations between the human body and skeleton. In the last stage, the weight coefficient of all terms keep certain, and the dataset is expanded to full scale. Using the output of the network as the initial value, the optimization with only one iteration is applied to provide regression supervision.

\begin{figure}[t]
\centering
\includegraphics[width=0.9\linewidth]{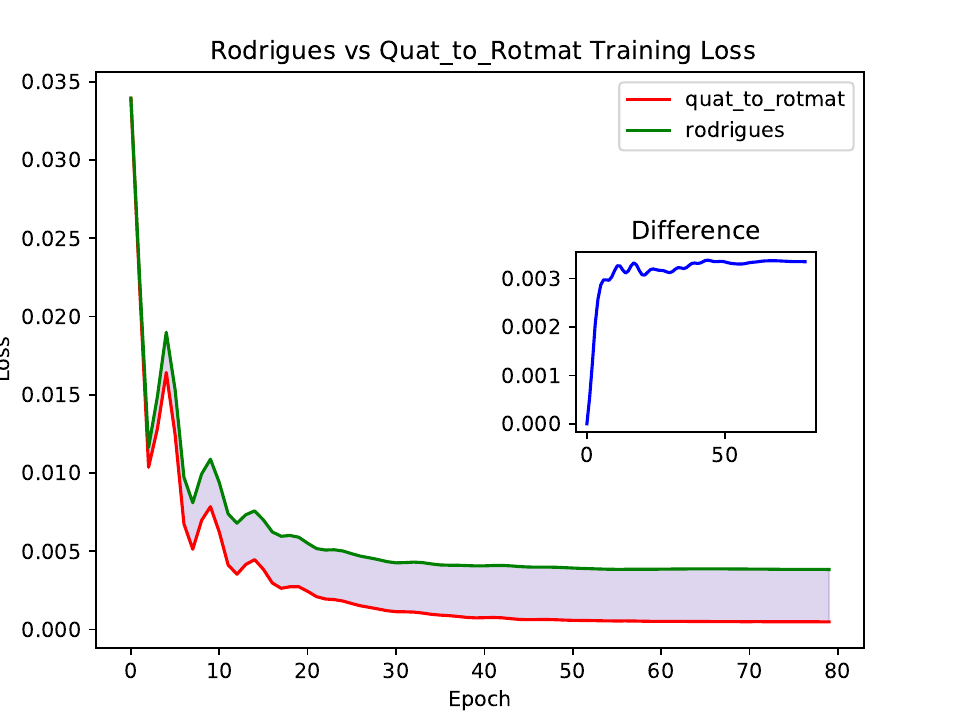}
\caption{The loss of \emph{Rodrigues} and \emph{Quaternion}}
\label{rodriloss}
\end{figure}

\textbf{Implementation Trick:} 
Compared with the \emph{Rodrigues} formula, we use \emph{Quaternion} method as an intermediary to obtain the rotation matrix.
Although choosing between two candidates based on
minimum error will always yield the same answer, these different process do affect the gradients of the search space and can yield slightly different results in practice (\ref{rodriloss}). Similar discussions can be found in \cite{7363472}.

\section{Experiments}

\subsection{Experiment Setup}

\textbf{Datasets:} We use a small-scale LSP \cite{johnson2010clustered} \emph{skin} dataset to train the network. To evaluate the effectiveness of the proposed \emph{skeletal} self-supervised training method, we also tested it on the large-scale 3DPW\cite{von2018recovering} dataset.

\textbf{Training details:}  Our three-stage progressive approach involves updating parameters in the Equation~\ref{skeleton_regression_format} during training. In the position demonstration stage (stage 1), our model is firstly trained on the 50\% LSP \cite{johnson2010clustered} for 500 epochs. In this state, we set $\lambda_{l}= 0$ , $\lambda_{r} = 0.1$, and $\lambda_{t}$ is set to $10$ to adjust the model learning of skeleton locations and features. In the weak supervision stage (stage 2), we increase $\lambda_{l}$ to 10, proving that the network can understand the correlation between skeletons in the training stage and ensure the connectivity between skeletons. In the collaborative improvement stage (stage 3), we restore $\lambda_{l} = 5$, $\lambda_{r} = 0.1$, $\lambda_{t} = 10$, and perform training on whole LSP to ensure the generalization of the network and avoid overfitting. Optimizer is set to Adam, and the initial learning rate is set to 1e-5, which is attenuated 0.9 times every 1500 steps.

\textbf{Experimental platform:} Experiments are performed on a single workstation with 4.20 GHz Intel I7 CPU and NVIDIA 2080.

\begin{figure}[ht]
\centering
\includegraphics[scale=0.5]{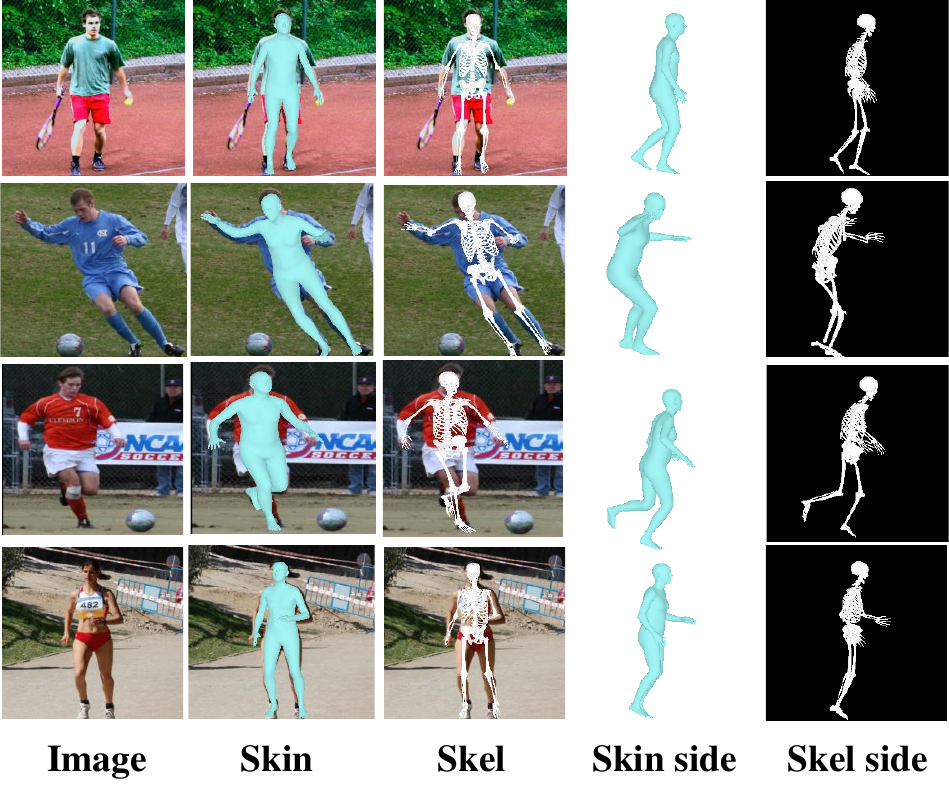}
\caption{Examples of EA-RAS on the LSP \cite{johnson2010clustered} test dataset. Other views of the human body and skeleton are also given.}
\label{vis2}
\end{figure}

\textbf{Evaluation metrics:} The performance of our method on datasets is shown in Figure~\ref{vis2}. In the following section, we quantitatively analyze skeleton reconstruction using metrics such as computation time, reconstruction error, and \emph{$D_{mean}$} (Euclidean Distance). For the generated human body skin, we evaluate it using common Per Vertex Error (PVE)\cite{pavllo20193d} and  Mean Per Joint Position Error (MPJPE)\cite{ionescu2013human3} metrics for comparison. Additionally, we analyze the time consumption of our proposed iterative optimization method. Moreover, we qualitatively evaluate some representative examples and share the results.

\subsection{Comparison with the Relevant Methods}
\begin{table*}[ht]

\caption{Skeleton results comparison on 3DPW\cite{von2018recovering} dataset.}

\centering
\resizebox{0.9\linewidth}{!}{
      \begin{tabular}{c|cccccc} 
    \toprule[1.5 pt]
    &Method & Human Mesh Source& gt-skeleton &  Skeleton Reconstruction Error$\downarrow $  & Skeleton $D_{mean}$ $\downarrow $  & Total Time$\downarrow $\\
    \midrule
    \midrule
    \multirow{7}*{\rotatebox{90}{3DPW}} 
    
    &OSSO \cite{Keller:CVPR:2022} CVPR'22&ROMP \cite{sun2021monocular} & 3DPW &86.5 &7.363 & 173.0\\
    &OSSO \cite{Keller:CVPR:2022} CVPR'22& BEV\cite{sun2022putting} & 3DPW &73.1 &3.532 & 176.5\\
    &OSSO \cite{Keller:CVPR:2022} CVPR'22& TRACE\cite{sun2023trace} & 3DPW & 83.8&1.015 & 183.6\\
    \cline{2-7}

    &EA-RAS& - & 3DPW & 83.2& 1.009 & \textbf{0.2}\\
    &EA-RAS+OSF$^+$& EA-RAS & 3DPW& 68.1 & 0.681 &  5.2\\
    &EA-RAS+PLUS& EA-RAS & 3DPW& \textbf{34.7} & \textbf{0.449} & 23.0\\

    \bottomrule[1.5 pt]
    \end{tabular}
    }
\label{skeleton-comp-3dpw}
\end{table*}

\subsubsection{Skeleton Reconstruction Result}

To test our approach, we use the human body skin mesh from the 3DPW dataset\cite{von2018recovering}, and obtained the skeleton ground truth using the same method as OSSO\cite{Keller:CVPR:2022}.
We conduct the experiment with two groups: one group uses a \emph{multi-stage} skeleton reconstruction method with OSSO\cite{Keller:CVPR:2022} completing the bone generation part and different methods\cite{sun2021monocular,sun2022putting,sun2023trace}  for skin reconstruction, while the other group uses our \emph{single-stage} method EA-RAS with optional subsequent optimization. As shown in Table~\ref{skeleton-comp-3dpw}, our method produces regression results of 83.2mm, 68.1mm with OSF$^+$, and 34.7mm with the OSF full version (PLUS). 
EA-RAS achieves comparable accuracy to \emph{multi-stage} based methods but significantly outperforms them in speed, completing the task in only 0.2 seconds compared to 173.0 seconds, over 800 times faster. Our OSF$^+$ and PLUS versions significantly reduced reconstruction errors by up to 34.7mm compared to the previous 73.1mm, representing a more than 50\% improvement. Additionally, the speed increased by over 7 times (173.0 / 23.0). The experimental results show that our method significantly improves the speed to meet real-time requirements. For scenarios with low real-time demands, our method greatly enhances accuracy while still ensuring effective speed improvement.

\begin{table}[ht]
\caption{Skeleton results on LSP\cite{johnson2010clustered} dataset.}
\centering
\resizebox{\linewidth}{!}{
      \begin{tabular}{c|ccc} 
    \toprule[1.5 pt]
    Method &   Reconstruction Error$\downarrow $  &  $ D_{mean}$
 $\downarrow $  & Total Time$\downarrow $\\
    \midrule
    \midrule
    OSSO \cite{Keller:CVPR:2022}&  52.9 &0.771& 140.8\\

    EA-RAS &68.9 & 0.734& \textbf{0.2}\\
    EA-RAS+OSF$^+$ &49.6 & 0.509 &  5.2\\
    EA-RAS+PLUS &\textbf{6.4} & \textbf{0.066}& 22.8\\
    \bottomrule[1.5 pt]
    \end{tabular}
    }

\label{skeleton-comp-lsp}
\end{table}

To further demonstrate our method, we also conduct skeletal comparison experiments on the LSP dataset. As the LSP\cite{johnson2010clustered} dataset does not contain human body mesh, we annotate skeletal point regression using EA-RAS's human body mesh to obtain ground truth. For fairness, we utilize the human body mesh obtained from EA-RAS for the mesh input required by OSSO\cite{Keller:CVPR:2022}. Results is reported in TABLE~\ref{skeleton-comp-lsp}. It can be observed that, with minimal difference in regression results, the introduction of our designed OSF$^+$ resulted in a reconstruction error of 49.6mm (vs. 52.9mm) within 5.2 seconds (vs. 140.8 seconds). Our PLUS version has achieved a reconstruction error of 6.4mm (vs. 52.9mm) within 22.8s (vs. 140.8s). Our optimization phase effectively integrates and guides the network, constraining and defining the search space during the learning process, resulting in closer outcomes. The method effectively combines the network and optimization.

\begin{figure}[ht]
\centering
\includegraphics[width=0.9\linewidth]{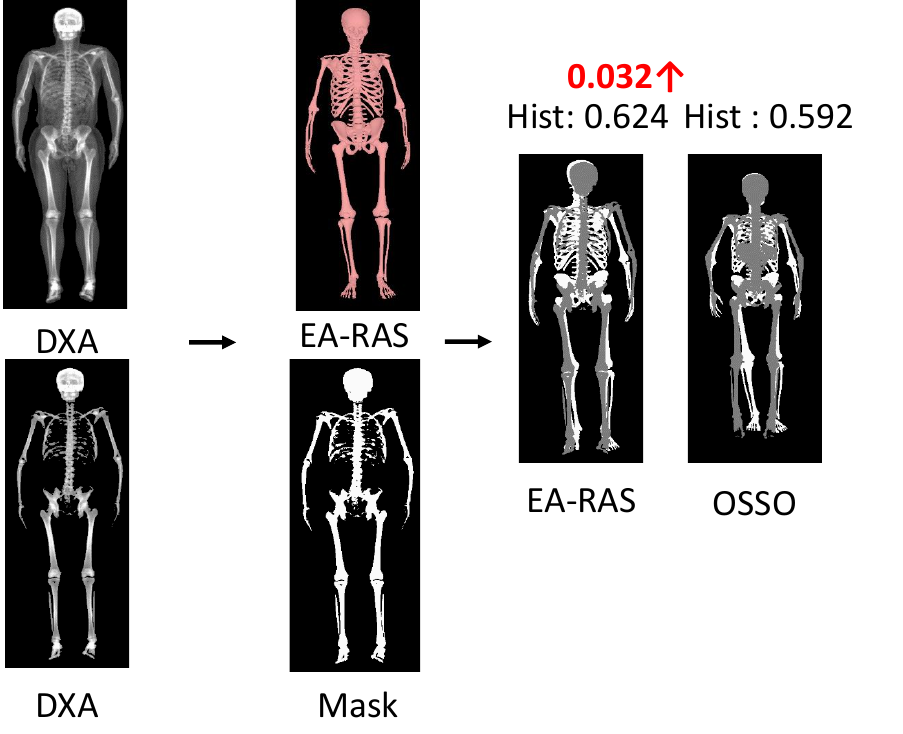}
\caption{Real bone results of EA-RAS. On the left is bone density map, a DXA image. Pink shows the skeleton result. The second picture below displays the bone mask from the density map. On the right, we projected bone results of OSSO and EA-RAS on the mask (gray is bone groundtruth). Similarity matching results and the groundtruth histogram is shown.}
\label{DXA}
\end{figure}

To verify the difference with real skeleton, we verified on a bone density data. As shown in Figure~\ref{DXA}, the model input is the DXA image of the human body (as shown in the image at the upper left corner of the figure). 

We control the threshold for the DXA image and obtain the bone density map. Because the DXA image is human anatomy illustration, it cannot clearly present the external human body. This error cause the OSSO algorithm to deviate from the correct skeleton because it relying on a correct human mesh in advance. Although the real bone density map cannot fully show the external contour of the bone subject to the physical occlusion of the machine, our output results are still 0.03 higher than the OSSO output and mask similarity.

\begin{figure}[ht]
\centering
\includegraphics[scale=0.6]{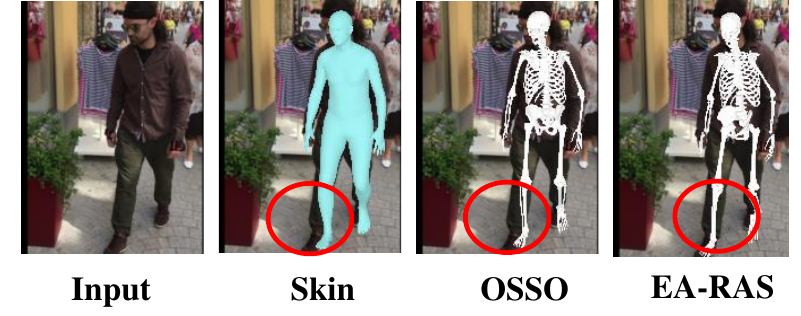}
\caption{Qualitative analysis results for OSSO and EA-RAS methods. Our model is derived from images and is not completely subject to inaccurate human body predictions avoiding the cumulative error caused by human predictions.}
\label{OIO}
\end{figure}

The skeleton model of the EA-RAS is directly deduced from the original image, in contrast to the OSSO model, which is generated based on the outcomes of human body reconstruction. As depicted in Figure~\ref{OIO}, our model aligns more closely with the original image, whereas the OSSO model relies on the human body mesh. This reliance could potentially lead to the introduction of extra errors

\subsubsection{Lightweight Skeleton Optimization}

As described in the Methodology section, we aim to develop a lightweight self-supervised approach for training regression skeletal parameters. We improve lightweight and optimization strategy to enhancing the speed.

The following comparison versions are set:

\begin{itemize}
\item OSSO \cite{Keller:CVPR:2022}: The original OSSO method has three steps: register body mesh point cloud to STAR model, infer the skeleton shape of lying pose and optimize the skeleton to the input pose.
\item OSSO* (Ours): To speed up the OSSO process, we remove step-3 in OSSO, just fit the input mesh to the STAR model with the corresponding pose directly, and then infer the skeleton.
\item OSSO** (Ours): To avoid unconstrained shift of skeleton position, the necessary migratory puppet \cite{zuffi2015stitched} loss mentioned in OSSO step-3 and clavicle joint control was added to OSSO*.
\item OSF (Ours): We skipped the process of registering STAR parameters from human mesh in OSSO**, and directly used the model parameters obtained by network regression and passed through transition.
\item OSF$^{+}$ (Ours): It is once option to improve accuracy and speed. It removes supervision of the last three terms in Formula \ref{skeleton_optimize_formula} and optimizes only once after the regression. 

\end{itemize}

\begin{table}[ht]
\caption{The comparison of time consumption between our improved optimization method and OSSO \cite{Keller:CVPR:2022} at various stages. The iteration numbers are chosen to a suitable value considering both the effect and speed. The unit is \emph{second / iteration}.}
\centering
    \resizebox{\linewidth}{!}{
    \begin{tabular}{ccccc} 
    \toprule[1.5 pt]

    \multirow{2}*{Method} & \multicolumn{4}{|c|}{Time Consumption (\emph{second / iteration})} \\
    \cline{2-5}
    ~ & \multicolumn{1}{|c|}{Register}  & \multicolumn{1}{c|}{Infer skeleton} & \multicolumn{1}{c|}{Pose skeleton} &
    \multicolumn{1}{c|}{Total} \\
    \midrule
    \midrule
    \multicolumn{1}{c|}{OSSO}  & 23.3 / 5  & 19.0 / 20  & 98.3 / (10+15)  &  140.6\\
    \multicolumn{1}{c|}{OSSO*}  & 35.9 / 7  & 34.1 / 40 & - & 70.0  \\
    \multicolumn{1}{c|}{OSSO**}  & 35.0 / 7  & 76.5 / 40  & - & 111.5 \\ 
 \multicolumn{1}{c|}{OSF} & -&32.3 / 15 
&-&32.3
\\
 \multicolumn{1}{c|}{OSF$^{+}$} & -&5.0 / 1 
&-&5.0
\\
    \bottomrule[1.5 pt]
    \end{tabular}}

\label{speedosso}
\end{table}

Table~\ref{speedosso} shows the comparison of time and iteration numbers among different methods to achieve similar results. 
Our method significantly improves the computational speed.
Compared with the original OSSO algorithm, our proposed improved version achieves a speed increase of more than 700 times (140.8 / 0.2), which can significantly save the time consumption with the minimum accuracy error, and can be applied in the self-supervised training process on the network.

\begin{figure}[ht]
\centering
\includegraphics[width=0.8\linewidth]{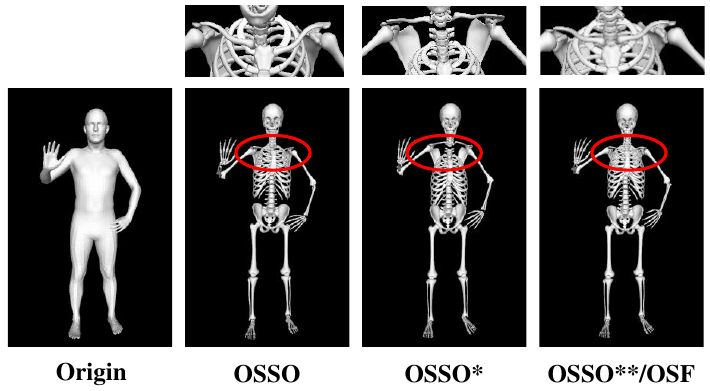}
\caption{The qualitative analysis of the improved optimization method and the OSSO method. Our improvements did not spoil OSSO's results.}
\label{skelquali}
\end{figure}

\begin{figure}[ht]
\centering
\includegraphics[width=\linewidth]{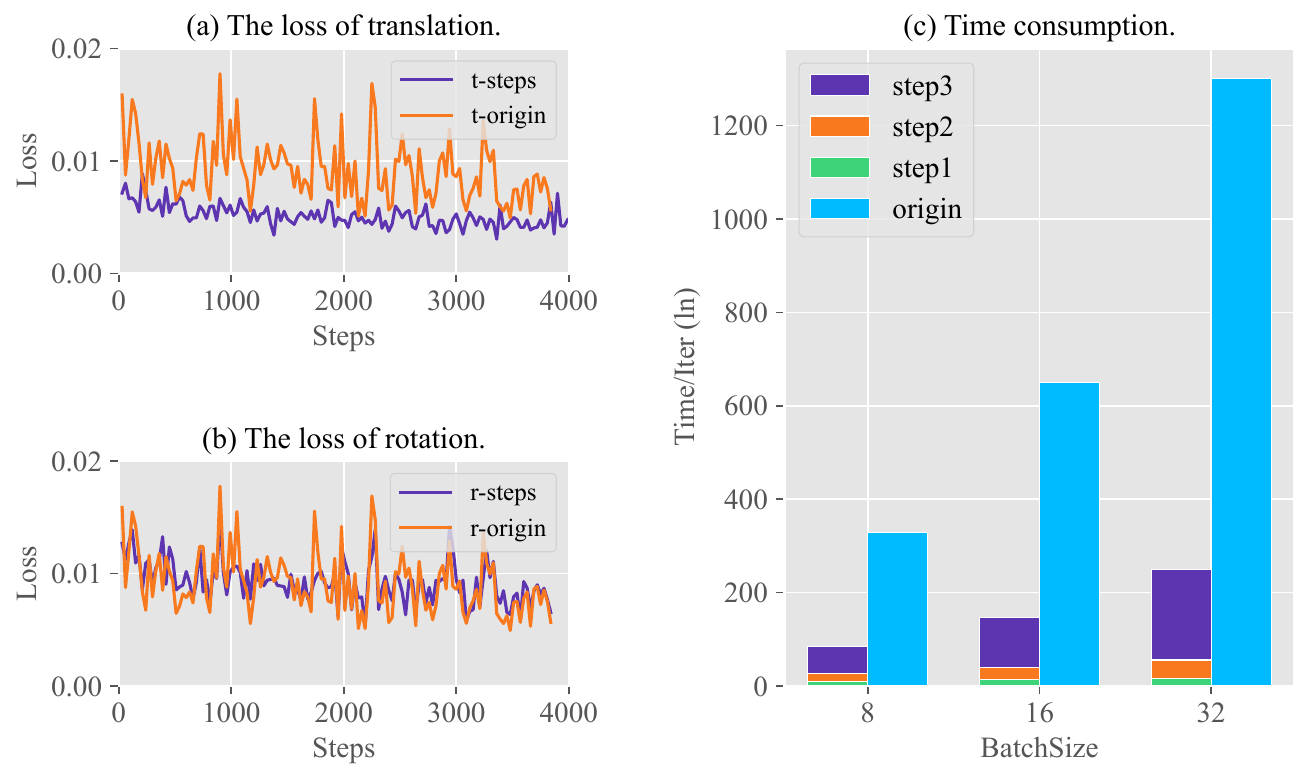}
\caption{The comparison of the effects of progressive training(steps) and directly training(origin). (a) is the translation loss of the skeleton, and (b) means rotation loss. The orange line represents our step-by-step progressive training method, and the purple represents the direct training method using the ground truth. (c) Quantitative results of the training duration of our various stages. Due to our optimization and improvement, the training speed of the model has been significantly improved compared to the direct training method.}
\label{40loss}
\end{figure}

As shown in Figure~\ref{skelquali}, we can also find the abnormal offset of clavicular joint. The deep learning method is used to learn the bone connection relationship, so that the optimization process can play an auxiliary role in it, which can greatly reduce this problem, and improve the original method. After adding the extra clavicle joint anatomical constraints in Formula ~\ref{skeleton_optimize_formula}, the skeleton result is better than the original OSSO and not affected by the increase in speed.

\subsubsection{Human Reconstruction Result}

\begin{table}[ht]

    \caption{ Reconstruction errors on 3DPW \cite{von2018recovering} testing set. Without using 3DPW training set, our model still achieved good results benefiting from the contribution of internal skeletal features. Bold indicates the maximum value, followed by a single underscore and double underscore.} 

\centering
\resizebox{\linewidth}{!}{
      \begin{tabular}{c|cccccc} 
    \toprule[1.5 pt]
    &Method & Intermediate & PVE$\downarrow $ & MPJPE$\downarrow$\\
    \midrule
    \midrule
    \multirow{8}*{\rotatebox{90}{Temporal}} 
    &HMMR \cite{kanazawa2019learning} CVPR'19&-& 139.3 &116.5\\
    &VIBE \cite{kocabas2020vibe} CVPR'20&- & 113.4 & 93.5\\
    &TCMR \cite{choi2021beyond} CVPR'21&-&111.3 & 95.0\\

    &MPS-Net \cite{wei2022capturing} CVPR'22&3D skeleton  &109.6 & 91.6\\
    &SHAPY \cite{choutas2022accurate} CVPR'22 &-&-&95.2\\
    &TePose \cite{wang2022tepose} TPAMI'22 & - & 115.9  & 93.9\\
    &CycleAdapt \cite{nam2023cyclic} ICCV'23&3D vertices & 107.0  &90.3\\

    \midrule
    \multirow{8}*{\rotatebox{90}{Frame-based}} 
    &HMR \cite{kanazawa2018end} CVPR'18&-&-  & 130.0\\
    &SPIN \cite{kolotouros2019learning} ICCV'19& 2D skeleton &  116.4  & 96.9\\
    &I2L-MeshNet \cite{moon2020i2l} ECCV'20&3D vertices& -  & 100.0\\
    &PyMAF \cite{zhang2021pymaf} ICCV'21& IUV image& 110.1  & 92.8\\
    &ROMP \cite{sun2021monocular} ICCV’21 &- &108.3  &91.3 \\
    &SmoothNet \cite{zeng2022smoothnet} ECCV'22 &- & 111.5 &97.8 \\
    & GLoT\cite{shen2023global} CVPR'23 &3D skeleton &107.8   &\textbf{89.9}   \\

    \cline{1-5}
    &\textbf{Ours}&2D skeleton &  \textbf{106.4}  &  91.7\\ 

    \bottomrule[1.5 pt]
    \end{tabular}
    }

\label{123}
\end{table}

Skeleton is generally built into the human body and satisfy anatomical constraints. During the experiment, we also found that the learning process of the skeleton is also refining and correcting the learning process of the human body. 

Table~\ref{123} shows result comparison on 3DPW dataset~\cite{von2018recovering}. 
It is noteworthy that our model is trained exclusively on a modest subset of datasets, deliberately excluding the 3DPW training set. Despite this limitation, when compared with state-of-the-art models that leverage the full 3DPW training set, our approach continues to yield good results in the realm of 3D human reconstruction.

\subsection{Ablation Study}
\subsubsection{Progressive training method}
For the proposed three-stage progressive training method, we compare it with the direct training method (\emph{i.e.} supervised the network with ground truth during the whole stage). Figure~\ref{40loss} displays the time consumption for ablation comparison experiments. We examine the effect of varying batch sizes on speed and depicted the loss convergence results in the same figure.

By gradually improving the network results step by step in a forward iterative manner, we find the process can lead the network convergence quickly and stably, while the direct data-driven method may lead to the unstable training effect. The experimental results show that the proposed progressive training process helps the network gradually learn information, reduces the complexity of the search process, and avoids the overfitting. 

\subsubsection{Extra Optimization Performance}

\begin{figure}[ht]
\centering
\includegraphics[width=\linewidth]{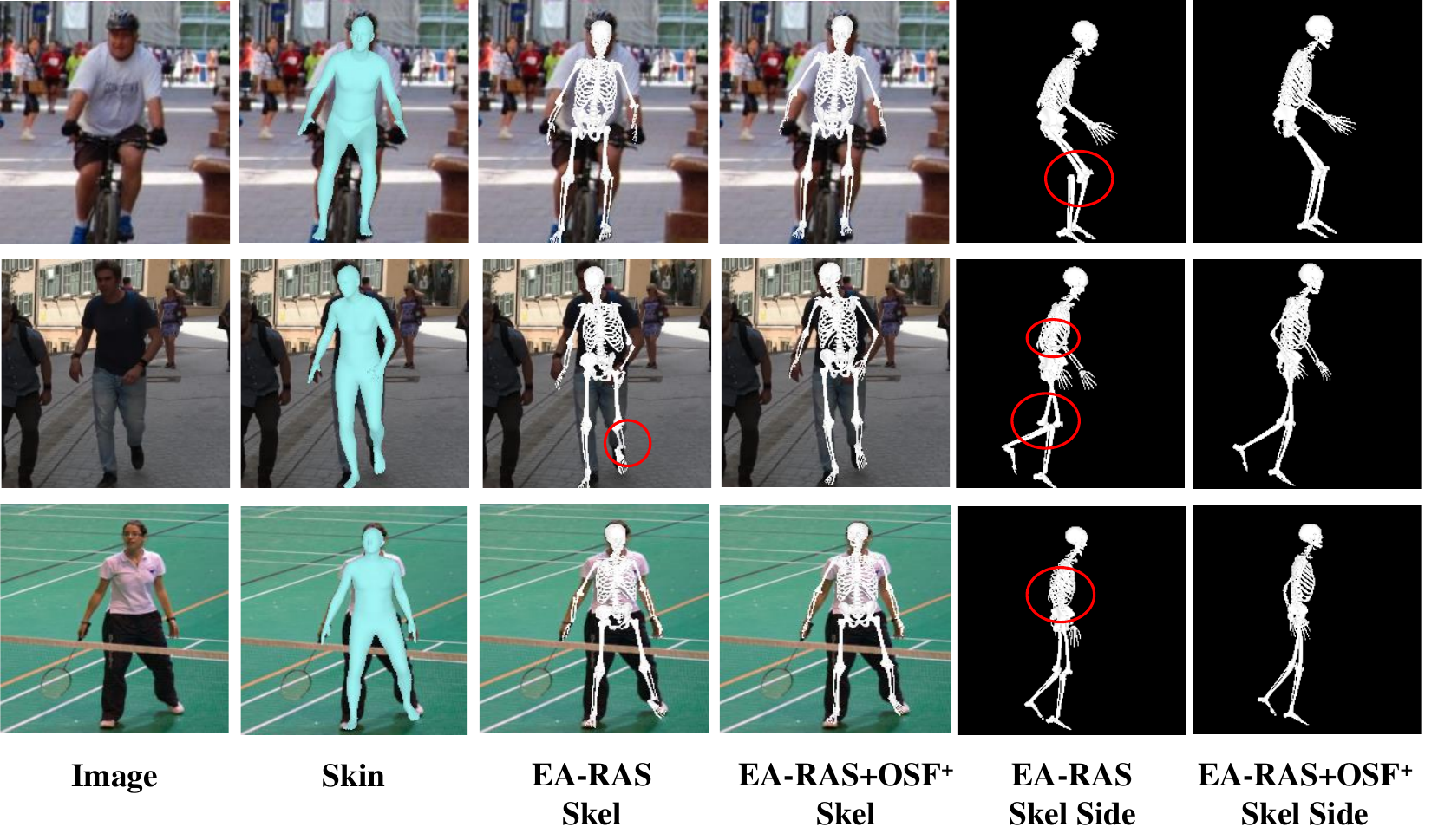}
\caption{On the dataset, the performance of EA-RAS and EA-RAS with once OSF$^{+}$ process. Adding an optimization helps to enhance the correlation of the human body to the skeleton.}
\label{vis1}
\end{figure}

Given that the input image provides solely two-dimensional information, the skeletal connections are inherently decoupled. Minor visual artifacts can lead to significant distortions in the bone alignment during complex movements. Consequently, augmenting the optimization process with enhanced constraints between bones is crucial for mitigating these distortions and ensuring anatomical accuracy.
Meanwhile, we show the performance of EA-RAS and EA-RAS+OSF$^{+}$ on the dataset (Figure~\ref{vis1}). Adding an extra optimization process reduces the anatomical relationship inference error and enhances the connection between the skeleton and the human body. 
The multi-view display highlights the model's performance ability in the direction. The model is still effective for the longitudinal inference of the image in the two-dimensional state.

 \begin{table}[ht]

    \caption{ Reconstruction errors on 3DPW \cite{von2018recovering} testing set. Without using 3DPW training set, our model still achieved good results benefiting from the contribution of internal skeletal features. Bold indicates the maximum value, followed by a single underscore and double underscore. The results, accompanied by additional optimization (KITRO\cite{yang2024kitro}, Smplify\cite{bogo2016keep}), are also presented.}

\centering
\resizebox{0.8\linewidth}{!}{
      \begin{tabular}{cccccc} 
    \toprule[1.5 pt]
     Skeleton-Way & optimation  & PVE$\downarrow $  & MPJPE$\downarrow $\\
    \midrule
    \midrule

     $\times$ & $\times$ &  116.4 &   96.9\\
     $\checkmark$ &$\times$&   \textbf{106.4}&     \textbf{91.7}\\
    \midrule
     $\times$ & Smplify  &  102.1  & 87.0\\ 
     $\checkmark$ &Smplify&  \textbf{92.4} &     \textbf{84.2}\\ 
    \midrule
     $\times$ & KITRO &  80.2  & 67.1\\
     $\checkmark$ & KITRO &\textbf{62.7}    &\textbf{53.5}\\ 

    \bottomrule[1.5 pt]
    \end{tabular}
    }

\label{abd_human_del_skel}
\end{table}

\subsubsection{Skeleton Helps Human Reconstruction}

As previously discussed, the reconstruction of anatomical skeletons necessitates the integration of skin surface data. In this section, we delve into the potential of the skeleton reconstruction process to concurrently enhance the precision of human body skin reconstruction. To this end, we conduct ablation studies utilizing the 3DPW dataset \cite{von2018recovering}, thereby assessing the interplay between skeletal and skin reconstruction accuracy.

Table~\ref{abd_human_del_skel} show the results of three sets of experiments. The first set involved removing the skeletal branch from the original model and predicting human and camera parameters directly. In the other two sets, we assessed if the skeletal branch could offer improved initial values for different human body optimization algorithm.
The results show that the branch with skeletal reconstruction can achieve better skin reconstruction effects. This suggests that in our model, the intrinsic information represented by the skeleton and the extrinsic information represented by the skin has been effectively fused.

\begin{figure}[htbp]
\centering
\includegraphics[width=0.9\linewidth]{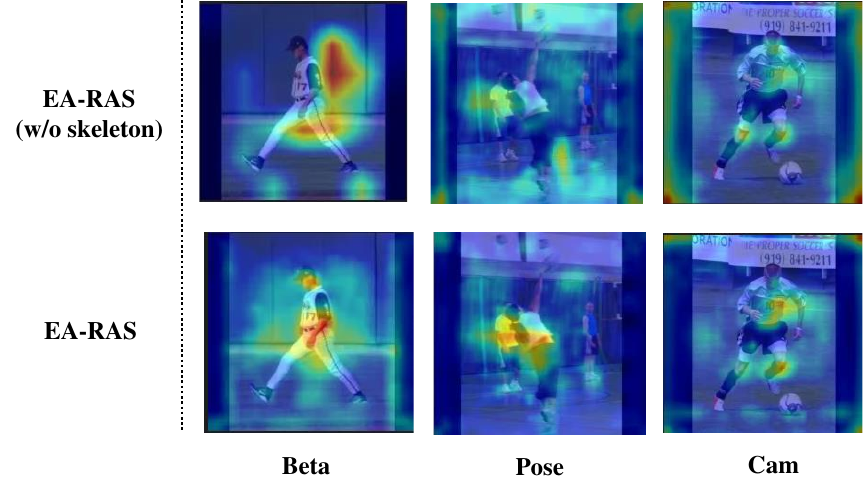}
\caption{The presentation of channel attention comparing the method without the skeletons inferring. The inputs are all low-definition pictures on LSP \cite{johnson2010clustered}. The attention map of $\beta$, \emph{pose}, \emph{cam} of the human body model, respectively. After the designed HTS, the model focuses more on the feature positions of the human body.}
\label{intro}
\end{figure}

Benefit from the anatomical position of the bones, the process of extracting bone features helps to refine and decompose human features. This process plays a crucial role in human prediction.
Figure~\ref{intro} shows the attention map of $\beta$, $pose$, and $cam$ parameters and find that EA-RAS is more focused on the position of the human compared with the case of reconstruction of branch roads by a single human body.

\subsection{Visualization of Training and Optimization Process}
\begin{figure}[ht]
\centering
\includegraphics[width=\linewidth]{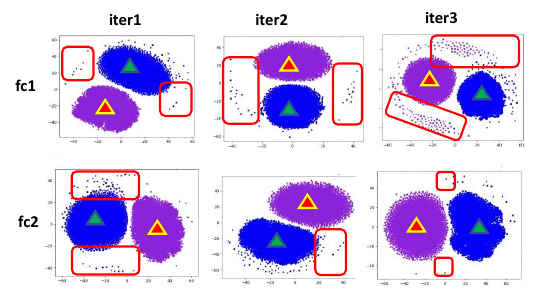}
\caption{Performance of TSNE \cite{van2008visualizing} features on the hrpet \cite{johnson2011learning} dataset under our HTS module. Human and skeletal features are well differentiated and fused.}
\label{htsksh}
\end{figure}

The human body and skeleton output are decoupled structures. While ensuring the correlation, our HTS has well integrated and differentiated them. Therefore, we have conducted dimension reduction feature visualization of the results.
As shown in Figure~\ref{htsksh}, we visualize the features in a reduced dimension. Blue areas represent skeletal features, while purple represents the human body. The triangle represents the cluster center, and iter1 to iter3 introduces the human body parameters into the skeleton feature vectors every time in Formula~\ref{HTS-skel-formula}. It can be seen that each time we perform HTS iteration, it bring the shocks to the feature classification (can be shown in the red box). Moreover, the feature distribution tends to converge with the forward propagation of parameters (from fc1 to fc2), which more significantly impacts the clustering. In the end, it forms two clusters, indicating that the skeleton and human body are strongly correlated but not completely overlapping. HTS helps the model to decouple and output features independently in the process of merging mutual features.

\begin{figure}[htbp]
\centering
\includegraphics[width=0.9\linewidth]{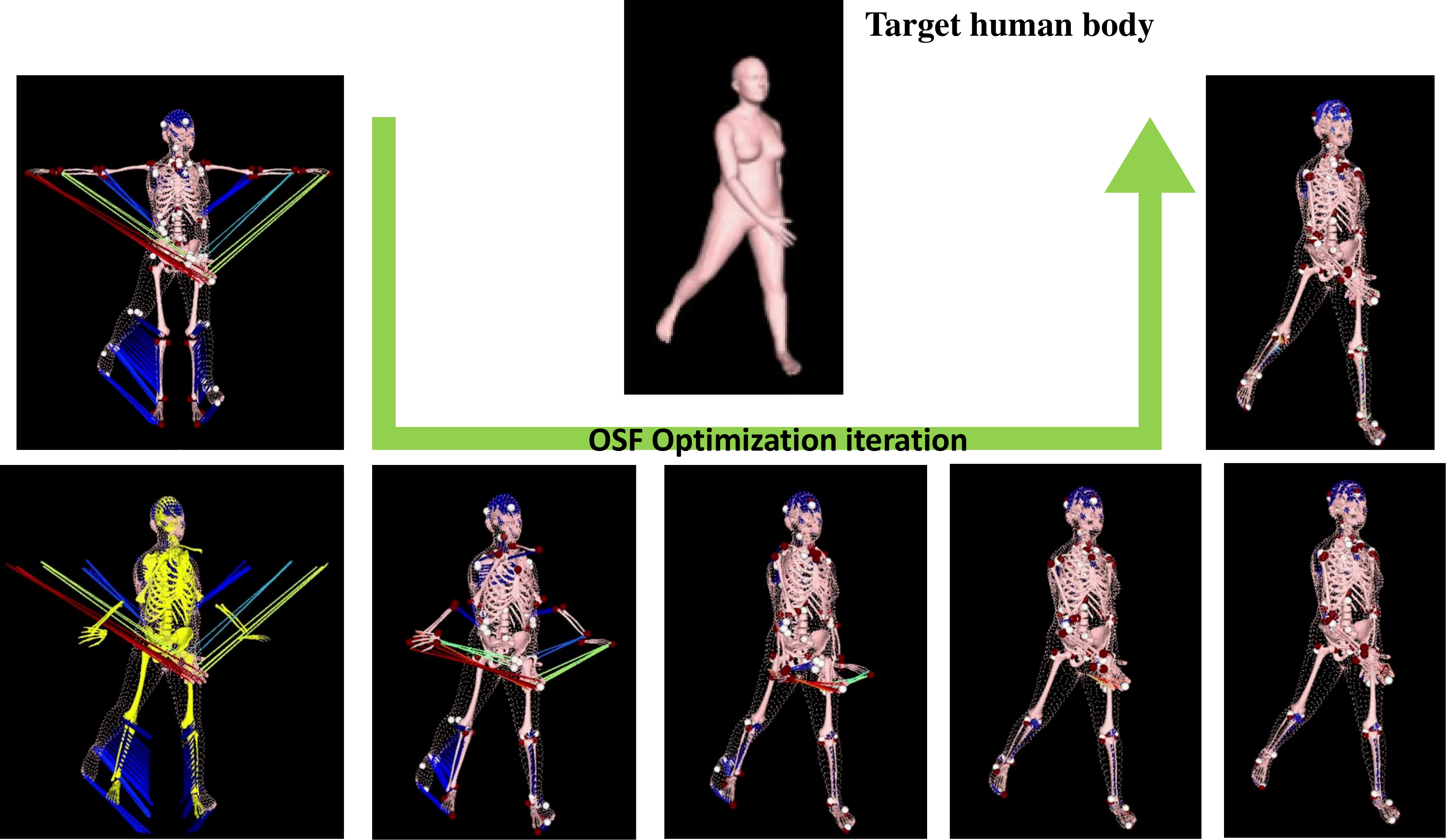}
\caption{It shows that the loss of OSF actually acts on the skeleton. The green arrows show the bone structure with optimization. Each color represents the loss in different areas, and the length shows the size of the loss. The picture in the middle shows the input external skin of the human body.}
\label{1231231}
\end{figure}

 \begin{table*}[ht]
\caption{We measure the pixel errors of six key starting acupoints in the recognition process. A pixel error of 10 corresponds roughly to 1(±0.3)cm. We also provided information on the errors associated with the three experts whose data we collected for our dataset. The x and y columns in the table show the average error in the axis direction, while 'dist' represents the Euclidean distance.}
\centering
\resizebox{\linewidth}{!}{
     \label{compareacu}
      \begin{tabular}{c|ccccccc|c} 
    \hline
    & Method &Left Gao Huang$\downarrow $ & Left Zhi Shi$\downarrow $ & First Thoracic Vertebra$\downarrow $ &  Fourth Lumbar Vertebra$\downarrow $ & Right Feng Men$\downarrow $ & Right Gan Shu$\downarrow $ & Total$\downarrow $ \\
    \hline
    \multirow{2}*{\cite{hu2021novel}}
    &x &16.32& 15.96 & 16.30& 10.18 & 11.24 & 11.22 &13.54\\
    &y &26.77 & 31.43 & 17.72 & 29.79 & 17.97 & 26.34 & 25.00\\
    
    \hline
    \multirow{2}*{EA-RAS}
    &x &15.95  & 13.34 & 6.18 & 10.67 & 8.64 & 9.24& 10.67\\
     &y &18.65 & 27.95 & 13.02 & 28.61 & 14.75 & 24.13&21.18\\
    \hline

    \end{tabular}
}
\end{table*}

Figure~\ref{1231231} shows the role of our loss in the standard human bone in OSF. Bone loss draws the bone block closer to the corresponding position step by step, making the final result meet the anatomical constraints.

\subsection{Application and Portability Experiment}
\begin{figure}[ht]
\centering
\includegraphics[width=0.9\linewidth]{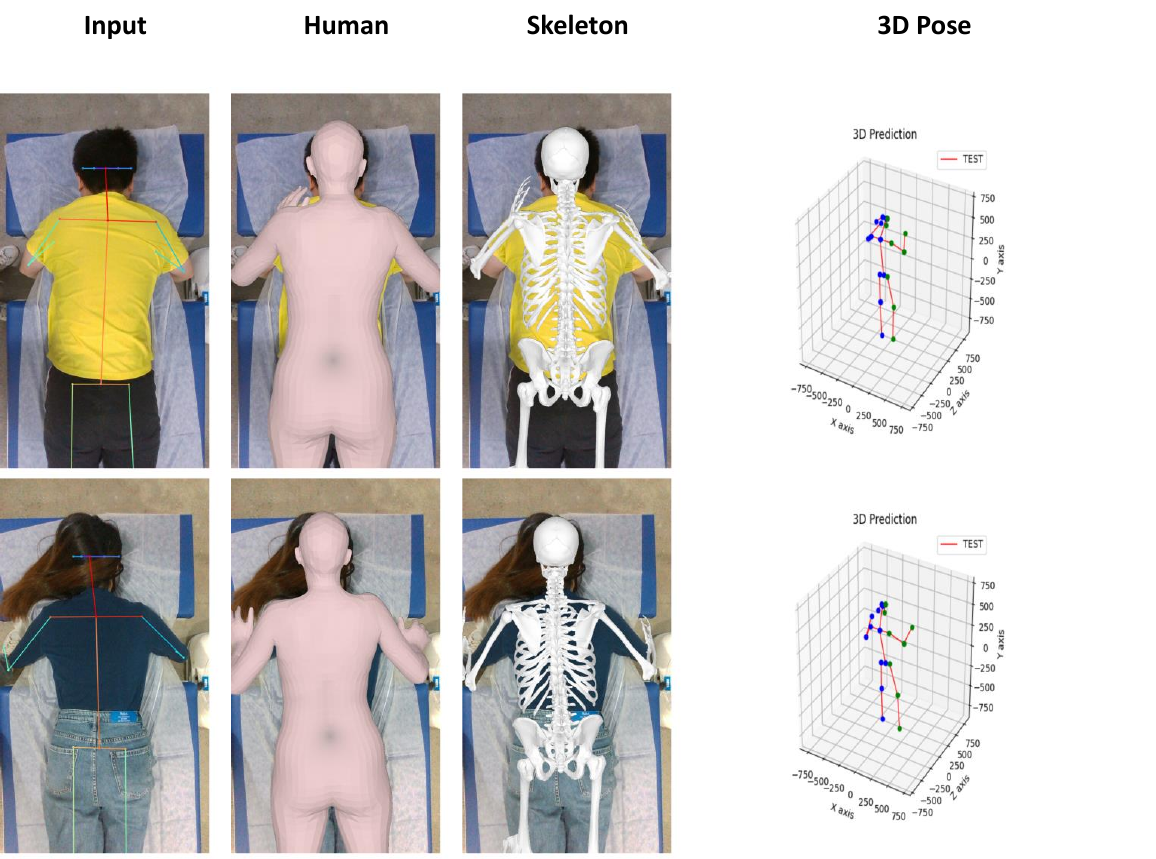}
\caption{The presentation of human body and skeleton in ADAT}
\label{vis2inputkey}
\end{figure}

To further validate the proposed method, we conduct user studies with a robot massage application \cite{hu2021novel,hu2013massage,li2020enhanced}. In this application, accurately locating the acupoints is a key requirement, since acupoints are typically located along the visceral meridians circulating around the skeleton and vertebrae in the human body\cite{acupointchineses,li2015acupoint}.
Precise identification of these acupoints is referred to as the Acupoint Recognition (OAR) problem. Previous methods\cite{9907739,9188367,hu2021novel} typically employed key point recognition algorithms to identify skeletal key points, thereby obtaining acupoint information. We view the OAR problem as a novel assessment of skeletal precision. For simplicity, we primarily evaluate the accuracy of the starting points of key acupoint meridian pathways.

To the best of our knowledge, there are no publicly available datasets for evaluating acupoint recognition algorithms. We collected a new Acupoint Dataset (ADAT) for algorithm performance evaluation. ADAT comprises 700 images of 100 participants captured using two different cameras, including images with various distorted postures. This dataset was annotated by experts from Zhejiang University School of Medicine in China who touched the participants' vertebrae.

We conduct relevant tests using EA-RAS+OSF$^{+}$. Results are presented at TABLE \ref{compareacu}. Compared to mainstream acupoint localization through key point methods, considering the size and scale relationship of bone nodes, our EA-RAS approach outperformed by 3-4 points on the ADAT dataset.% Moreover, we visualized the distribution of acupoints and qualitative results  at Fig.~\ref{vis2acu}. Horizontally, EA-RAS accounts for bone width, allowing for more precise acupoint positioning based on bone width. Vertically, we have refined information on bone joint lengths, enabling simultaneous determination of acupoint positions and adjacent bone positions.

The results demonstrate the versatility of EA-RAS, as it can be readily applied to bone-dependent localization problems, yielding more accurate and precise positioning within the same domain-specific knowledge. While still in its early stages with respect to widespread acupoint applications, EA-RAS offers a new solution for OAR and exhibits promising accuracy capabilities.

% \begin{figure}[ht]
% \centering
% \includegraphics[width=\linewidth]{img/exp_acup_distribute.png}
% \caption{The human acupoints results are shown in red for expert data, green for skeleton inference. (a) Examples of the datasets. (b) is the distribute of acupoints between the output of our model and ground truth.}
% \label{vis2acu}
% \end{figure}

\section{Conclusion}

In this paper, we propose a single-stage, lightweight, and plug-and-play anatomical skeleton estimator, which achieves real-time, accurate anatomically realistic skeleton reconstruction with arbitrary pose using only a single RGB image input. 
We also introduce a dual-branch mechanism to fuse skin information in the skeleton estimation process to avoid the information loss and inaccuracies that exists in multi-stage methods. 
The dual-branch mechanism considers the human anatomical constraints and forms a tight connection between the human body and the anatomical skeleton. 
Moreover, we propose a three-stage progressive training method to eliminates the need for complex full skeleton annotation and ensure fast convergence. our training method can achieve self-supervision subsequently through a enhanced optimization process. 
Experiments show that our method has significant advantages over the relevant methods in terms of speed and accuracy.

% \section*{Acknowledgments}
% This should be a simple paragraph before the References to thank those individuals and institutions who have supported your work on this article.

% {\appendix[Proof of the Zonklar Equations]
% Use $\backslash${\tt{appendix}} if you have a single appendix:
% Do not use $\backslash${\tt{section}} anymore after $\backslash${\tt{appendix}}, only $\backslash${\tt{section*}}.
% If you have multiple appendixes use $\backslash${\tt{appendices}} then use $\backslash${\tt{section}} to start each appendix.
% You must declare a $\backslash${\tt{section}} before using any $\backslash${\tt{subsection}} or using $\backslash${\tt{label}} ($\backslash${\tt{appendices}} by itself
%  starts a section numbered zero.)}

% \section{References Section}
% You can use a bibliography generated by BibTeX as a .bbl file.
%  BibTeX documentation can be easily obtained at:
%  http://mirror.ctan.org/biblio/bibtex/contrib/doc/
%  The IEEEtran BibTeX style support page is:
%  http://www.michaelshell.org/tex/ieeetran/bibtex/

% You can manually copy in the resultant .bbl file and set second argument of $\backslash${\tt{begin}} to the number of references
%  (used to reserve space for the reference number labels box).

\bibliographystyle{IEEEtran}
\bibliography{earas-ref}

\begin{thebibliography}{10}
\providecommand{\url}[1]{#1}
\csname url@samestyle\endcsname
\providecommand{\newblock}{\relax}
\providecommand{\bibinfo}[2]{#2}
\providecommand{\BIBentrySTDinterwordspacing}{\spaceskip=0pt\relax}
\providecommand{\BIBentryALTinterwordstretchfactor}{4}
\providecommand{\BIBentryALTinterwordspacing}{\spaceskip=\fontdimen2\font plus
\BIBentryALTinterwordstretchfactor\fontdimen3\font minus \fontdimen4\font\relax}
\providecommand{\BIBforeignlanguage}[2]{{%
\expandafter\ifx\csname l@#1\endcsname\relax
\typeout{** WARNING: IEEEtran.bst: No hyphenation pattern has been}%
\typeout{** loaded for the language `#1'. Using the pattern for}%
\typeout{** the default language instead.}%
\else
\language=\csname l@#1\endcsname
\fi
#2}}
\providecommand{\BIBdecl}{\relax}
\BIBdecl

\bibitem{balmik2022nao}
A.~Balmik, M.~Jha, and A.~Nandy, ``Nao robot teleoperation with human motion recognition,'' \emph{Arabian Journal for Science and Engineering}, vol.~47, no.~2, pp. 1137--1146, 2022.

\bibitem{liu2022spatial}
S.~Liu, W.~Wu, J.~Wu, and Y.~Lin, ``Spatial-temporal parallel transformer for arm-hand dynamic estimation,'' in \emph{Proceedings of the IEEE/CVF Conference on Computer Vision and Pattern Recognition}, 2022, pp. 20\,523--20\,532.

\bibitem{ijcai2021-0135}
\BIBentryALTinterwordspacing
H.~Rao, S.~Xu, X.~Hu, J.~Cheng, and B.~Hu, ``Multi-level graph encoding with structural-collaborative relation learning for skeleton-based person re-identification,'' in \emph{Proceedings of the Thirtieth International Joint Conference on Artificial Intelligence, {IJCAI-21}}, Z.-H. Zhou, Ed.\hskip 1em plus 0.5em minus 0.4em\relax International Joint Conferences on Artificial Intelligence Organization, 8 2021, pp. 973--980, main Track. [Online]. Available: \url{https://doi.org/10.24963/ijcai.2021/135}
\BIBentrySTDinterwordspacing

\bibitem{cao2017realtime}
Z.~Cao, T.~Simon, S.-E. Wei, and Y.~Sheikh, ``Realtime multi-person 2d pose estimation using part affinity fields,'' in \emph{Proceedings of the IEEE conference on computer vision and pattern recognition}, 2017, pp. 7291--7299.

\bibitem{xu2022vitpose}
Y.~Xu, J.~Zhang, Q.~Zhang, and D.~Tao, ``Vitpose: Simple vision transformer baselines for human pose estimation,'' \emph{Advances in Neural Information Processing Systems}, vol.~35, pp. 38\,571--38\,584, 2022.

\bibitem{sun2019deep}
K.~Sun, B.~Xiao, D.~Liu, and J.~Wang, ``Deep high-resolution representation learning for human pose estimation,'' in \emph{Proceedings of the IEEE/CVF conference on computer vision and pattern recognition}, 2019, pp. 5693--5703.

\bibitem{hu2021novel}
W.~Hu, Q.~Sheng, and X.~Sheng, ``A novel realtime vision-based acupoint estimation for tcm massage robot,'' in \emph{2021 27th International Conference on Mechatronics and Machine Vision in Practice (M2VIP)}.\hskip 1em plus 0.5em minus 0.4em\relax IEEE, 2021, pp. 771--776.

\bibitem{xu2024toward}
Q.~Xu, Z.~Deng, C.~Zeng, Z.~Li, B.~He, and J.~Zhang, ``Toward automatic robotic massage based on interactive trajectory planning and control,'' \emph{Complex \& Intelligent Systems}, vol.~10, no.~3, pp. 4397--4407, 2024.

\bibitem{stevens2024bioclip}
S.~Stevens, J.~Wu, M.~J. Thompson, E.~G. Campolongo, C.~H. Song, D.~E. Carlyn, L.~Dong, W.~M. Dahdul, C.~Stewart, T.~Berger-Wolf \emph{et~al.}, ``Bioclip: A vision foundation model for the tree of life,'' in \emph{Proceedings of the IEEE/CVF Conference on Computer Vision and Pattern Recognition}, 2024, pp. 19\,412--19\,424.

\bibitem{siyao2022bailando}
L.~Siyao, W.~Yu, T.~Gu, C.~Lin, Q.~Wang, C.~Qian, C.~C. Loy, and Z.~Liu, ``Bailando: 3d dance generation by actor-critic gpt with choreographic memory,'' in \emph{Proceedings of the IEEE/CVF Conference on Computer Vision and Pattern Recognition}, 2022, pp. 11\,050--11\,059.

\bibitem{saito2015computational}
S.~Saito, Z.-Y. Zhou, and L.~Kavan, ``Computational bodybuilding: Anatomically-based modeling of human bodies,'' \emph{ACM Transactions on Graphics (TOG)}, vol.~34, no.~4, pp. 1--12, 2015.

\bibitem{ali2013anatomy}
D.~Ali-Hamadi, T.~Liu, B.~Gilles, L.~Kavan, F.~Faure, O.~Palombi, and M.-P. Cani, ``Anatomy transfer,'' \emph{ACM transactions on graphics (TOG)}, vol.~32, no.~6, pp. 1--8, 2013.

\bibitem{Keller:CVPR:2022}
M.~Keller, S.~Zuffi, M.~J. Black, and S.~Pujades, ``{OSSO}: Obtaining skeletal shape from outside,'' in \emph{Proceedings IEEE/CVF Conf.~on Computer Vision and Pattern Recognition (CVPR)}, Jun. 2022, pp. 20\,492--20\,501.

\bibitem{kadlevcek2016reconstructing}
P.~Kadle{\v{c}}ek, A.-E. Ichim, T.~Liu, J.~K{\v{r}}iv{\'a}nek, and L.~Kavan, ``Reconstructing personalized anatomical models for physics-based body animation,'' \emph{ACM Transactions on Graphics (TOG)}, vol.~35, no.~6, pp. 1--13, 2016.

\bibitem{von2018recovering}
T.~Von~Marcard, R.~Henschel, M.~J. Black, B.~Rosenhahn, and G.~Pons-Moll, ``Recovering accurate 3d human pose in the wild using imus and a moving camera,'' in \emph{Proceedings of the European Conference on Computer Vision (ECCV)}, 2018, pp. 601--617.

\bibitem{johnson2010clustered}
S.~Johnson and M.~Everingham, ``Clustered pose and nonlinear appearance models for human pose estimation.'' in \emph{bmvc}.\hskip 1em plus 0.5em minus 0.4em\relax Aberystwyth, UK, 2010, p.~5.

\bibitem{zimmermann2017learning}
C.~Zimmermann and T.~Brox, ``Learning to estimate 3d hand pose from single rgb images,'' in \emph{Proceedings of the IEEE international conference on computer vision}, 2017, pp. 4903--4911.

\bibitem{xu2019denserac}
Y.~Xu, S.-C. Zhu, and T.~Tung, ``Denserac: Joint 3d pose and shape estimation by dense render-and-compare,'' in \emph{Proceedings of the IEEE/CVF International Conference on Computer Vision}, 2019, pp. 7760--7770.

\bibitem{guler2018densepose}
R.~A. G{\"u}ler, N.~Neverova, and I.~Kokkinos, ``Densepose: Dense human pose estimation in the wild,'' in \emph{Proceedings of the IEEE conference on computer vision and pattern recognition}, 2018, pp. 7297--7306.

\bibitem{loper2014mosh}
M.~Loper, N.~Mahmood, and M.~J. Black, ``Mosh: Motion and shape capture from sparse markers,'' \emph{ACM Transactions on Graphics (ToG)}, vol.~33, no.~6, pp. 1--13, 2014.

\bibitem{mahmood2019amass}
N.~Mahmood, N.~Ghorbani, N.~F. Troje, G.~Pons-Moll, and M.~J. Black, ``Amass: Archive of motion capture as surface shapes,'' in \emph{Proceedings of the IEEE/CVF international conference on computer vision}, 2019, pp. 5442--5451.

\bibitem{lassner2017unite}
C.~Lassner, J.~Romero, M.~Kiefel, F.~Bogo, M.~J. Black, and P.~V. Gehler, ``Unite the people: Closing the loop between 3d and 2d human representations,'' in \emph{Proceedings of the IEEE conference on computer vision and pattern recognition}, 2017, pp. 6050--6059.

\bibitem{kolotouros2019learning}
N.~Kolotouros, G.~Pavlakos, M.~J. Black, and K.~Daniilidis, ``Learning to reconstruct 3d human pose and shape via model-fitting in the loop,'' in \emph{Proceedings of the IEEE/CVF International Conference on Computer Vision}, 2019, pp. 2252--2261.

\bibitem{andriluka2009pictorial}
M.~Andriluka, S.~Roth, and B.~Schiele, ``Pictorial structures revisited: People detection and articulated pose estimation,'' in \emph{2009 IEEE conference on computer vision and pattern recognition}.\hskip 1em plus 0.5em minus 0.4em\relax IEEE, 2009, pp. 1014--1021.

\bibitem{gkioxari2013articulated}
G.~Gkioxari, P.~Arbel{\'a}ez, L.~Bourdev, and J.~Malik, ``Articulated pose estimation using discriminative armlet classifiers,'' in \emph{Proceedings of the IEEE conference on computer vision and pattern recognition}, 2013, pp. 3342--3349.

\bibitem{yi2022human}
H.~Yi, C.-H.~P. Huang, D.~Tzionas, M.~Kocabas, M.~Hassan, S.~Tang, J.~Thies, and M.~J. Black, ``Human-aware object placement for visual environment reconstruction,'' in \emph{Proceedings of the IEEE/CVF Conference on Computer Vision and Pattern Recognition}, 2022, pp. 3959--3970.

\bibitem{zou2021modulated}
Z.~Zou and W.~Tang, ``Modulated graph convolutional network for 3d human pose estimation,'' in \emph{Proceedings of the IEEE/CVF international conference on computer vision}, 2021, pp. 11\,477--11\,487.

\bibitem{zhang2021adafuse}
Z.~Zhang, C.~Wang, W.~Qiu, W.~Qin, and W.~Zeng, ``Adafuse: Adaptive multiview fusion for accurate human pose estimation in the wild,'' \emph{International Journal of Computer Vision}, vol. 129, pp. 703--718, 2021.

\bibitem{papandreou2017towards}
G.~Papandreou, T.~Zhu, N.~Kanazawa, A.~Toshev, J.~Tompson, C.~Bregler, and K.~Murphy, ``Towards accurate multi-person pose estimation in the wild,'' in \emph{Proceedings of the IEEE conference on computer vision and pattern recognition}, 2017, pp. 4903--4911.

\bibitem{he2017mask}
K.~He, G.~Gkioxari, P.~Doll{\'a}r, and R.~Girshick, ``Mask r-cnn,'' in \emph{Proceedings of the IEEE international conference on computer vision}, 2017, pp. 2961--2969.

\bibitem{newell2017associative}
A.~Newell, Z.~Huang, and J.~Deng, ``Associative embedding: End-to-end learning for joint detection and grouping,'' \emph{Advances in neural information processing systems}, vol.~30, 2017.

\bibitem{wang2018stacked}
D.~Wang, ``Stacked dense-hourglass networks for human pose estimation,'' Ph.D. dissertation, University of Illinois at Urbana-Champaign, 2018.

\bibitem{carreira2016human}
J.~Carreira, P.~Agrawal, K.~Fragkiadaki, and J.~Malik, ``Human pose estimation with iterative error feedback,'' in \emph{Proceedings of the IEEE conference on computer vision and pattern recognition}, 2016, pp. 4733--4742.

\bibitem{yang2021transpose}
S.~Yang, Z.~Quan, M.~Nie, and W.~Yang, ``Transpose: Keypoint localization via transformer,'' in \emph{Proceedings of the IEEE/CVF international conference on computer vision}, 2021, pp. 11\,802--11\,812.

\bibitem{li2021tokenpose}
Y.~Li, S.~Zhang, Z.~Wang, S.~Yang, W.~Yang, S.-T. Xia, and E.~Zhou, ``Tokenpose: Learning keypoint tokens for human pose estimation,'' in \emph{Proceedings of the IEEE/CVF International conference on computer vision}, 2021, pp. 11\,313--11\,322.

\bibitem{besl1992method}
P.~J. Besl and N.~D. McKay, ``Method for registration of 3-d shapes,'' in \emph{Sensor fusion IV: control paradigms and data structures}, vol. 1611.\hskip 1em plus 0.5em minus 0.4em\relax Spie, 1992, pp. 586--606.

\bibitem{bronstein2008numerical}
A.~M. Bronstein, M.~M. Bronstein, and R.~Kimmel, \emph{Numerical geometry of non-rigid shapes}.\hskip 1em plus 0.5em minus 0.4em\relax Springer Science \& Business Media, 2008.

\bibitem{mateus2008articulated}
D.~Mateus, R.~Horaud, D.~Knossow, F.~Cuzzolin, and E.~Boyer, ``Articulated shape matching using laplacian eigenfunctions and unsupervised point registration,'' in \emph{2008 IEEE Conference on Computer Vision and Pattern Recognition}.\hskip 1em plus 0.5em minus 0.4em\relax IEEE, 2008, pp. 1--8.

\bibitem{lipman2009mobius}
Y.~Lipman and T.~Funkhouser, ``M{\"o}bius voting for surface correspondence,'' \emph{ACM Transactions on Graphics (ToG)}, vol.~28, no.~3, pp. 1--12, 2009.

\bibitem{bauer2016modelisation}
A.~Bauer, ``Mod{\'e}lisation anatomique utilisateur-sp{\'e}cifique et animation temps-r{\'e}el: Application {\`a} l'apprentissage de l'anatomie,'' Ph.D. dissertation, Universit{\'e} Grenoble Alpes (ComUE), 2016.

\bibitem{balan2008naked}
A.~BALAN, ``The naked truth: Estimating body shape under clothing,'' \emph{ECCV, 2008}, 2008.

\bibitem{hasler2010multilinear}
N.~Hasler, H.~Ackermann, B.~Rosenhahn, T.~Thorm{\"a}hlen, and H.-P. Seidel, ``Multilinear pose and body shape estimation of dressed subjects from image sets,'' in \emph{CVPR}, 2010.

\bibitem{bogo2016keep}
F.~Bogo, A.~Kanazawa, C.~Lassner, P.~Gehler, J.~Romero, and M.~J. Black, ``Keep it smpl: Automatic estimation of 3d human pose and shape from a single image,'' in \emph{European conference on computer vision}.\hskip 1em plus 0.5em minus 0.4em\relax Springer, 2016, pp. 561--578.

\bibitem{8953618}
D.~Xiang, H.~Joo, and Y.~Sheikh, ``Monocular total capture: Posing face, body, and hands in the wild,'' in \emph{2019 IEEE/CVF Conference on Computer Vision and Pattern Recognition (CVPR)}, 2019, pp. 10\,957--10\,966.

\bibitem{9010321}
M.~Hassan, V.~Choutas, D.~Tzionas, and M.~Black, ``Resolving 3d human pose ambiguities with 3d scene constraints,'' in \emph{2019 IEEE/CVF International Conference on Computer Vision (ICCV)}, 2019, pp. 2282--2292.

\bibitem{10.1007/978-3-030-58610-2_3}
\BIBentryALTinterwordspacing
J.~Y. Zhang, S.~Pepose, H.~Joo, D.~Ramanan, J.~Malik, and A.~Kanazawa, ``Perceiving 3d human-object spatial arrangements from a single image in the wild,'' in \emph{Computer Vision – ECCV 2020: 16th European Conference, Glasgow, UK, August 23–28, 2020, Proceedings, Part XII}.\hskip 1em plus 0.5em minus 0.4em\relax Berlin, Heidelberg: Springer-Verlag, 2020, p. 34–51. [Online]. Available: \url{https://doi.org/10.1007/978-3-030-58610-2_3}
\BIBentrySTDinterwordspacing

\bibitem{kocabas2021pare}
M.~Kocabas, C.-H.~P. Huang, O.~Hilliges, and M.~J. Black, ``Pare: Part attention regressor for 3d human body estimation,'' in \emph{Proceedings of the IEEE/CVF International Conference on Computer Vision}, 2021, pp. 11\,127--11\,137.

\bibitem{choi2020pose2mesh}
H.~Choi, G.~Moon, and K.~M. Lee, ``Pose2mesh: Graph convolutional network for 3d human pose and mesh recovery from a 2d human pose,'' in \emph{European Conference on Computer Vision}.\hskip 1em plus 0.5em minus 0.4em\relax Springer, 2020, pp. 769--787.

\bibitem{hanocka2020point2mesh}
R.~Hanocka, G.~Metzer, R.~Giryes, and D.~Cohen-Or, ``Point2mesh: A self-prior for deformable meshes,'' \emph{arXiv preprint arXiv:2005.11084}, 2020.

\bibitem{varol2018bodynet}
G.~Varol, D.~Ceylan, B.~Russell, J.~Yang, E.~Yumer, I.~Laptev, and C.~Schmid, ``Bodynet: Volumetric inference of 3d human body shapes,'' in \emph{Proceedings of the European conference on computer vision (ECCV)}, 2018, pp. 20--36.

\bibitem{wang2018pixel2mesh}
N.~Wang, Y.~Zhang, Z.~Li, Y.~Fu, W.~Liu, and Y.-G. Jiang, ``Pixel2mesh: Generating 3d mesh models from single rgb images,'' in \emph{Proceedings of the European conference on computer vision (ECCV)}, 2018, pp. 52--67.

\bibitem{vaswani2017attention}
A.~Vaswani, N.~Shazeer, N.~Parmar, J.~Uszkoreit, L.~Jones, A.~N. Gomez, {\L}.~Kaiser, and I.~Polosukhin, ``Attention is all you need,'' \emph{Advances in neural information processing systems}, vol.~30, 2017.

\bibitem{lin2021end-to-end}
K.~Lin, L.~Wang, and Z.~Liu, ``End-to-end human pose and mesh reconstruction with transformers,'' in \emph{CVPR}, 2021.

\bibitem{lin2021-mesh-graphormer}
------, ``Mesh graphormer,'' in \emph{ICCV}, 2021.

\bibitem{zhang2021pymaf}
H.~Zhang, Y.~Tian, X.~Zhou, W.~Ouyang, Y.~Liu, L.~Wang, and Z.~Sun, ``Pymaf: 3d human pose and shape regression with pyramidal mesh alignment feedback loop,'' in \emph{Proceedings of the IEEE/CVF International Conference on Computer Vision}, 2021, pp. 11\,446--11\,456.

\bibitem{Pavlakos2018}
G.~Pavlakos, L.~Zhu, X.~Zhou, and K.~Daniilidis, ``Learning to estimate 3d human pose and shape from a single color image,'' in \emph{Proceedings of the IEEE conference on computer vision and pattern recognition}, 2018, pp. 459--468.

\bibitem{Omran2018}
M.~Omran, C.~Lassner, G.~Pons-Moll, P.~Gehler, and B.~Schiele, ``Neural body fitting: Unifying deep learning and model based human pose and shape estimation,'' in \emph{2018 international conference on 3D vision (3DV)}.\hskip 1em plus 0.5em minus 0.4em\relax IEEE, 2018, pp. 484--494.

\bibitem{kanazawa2018end}
A.~Kanazawa, M.~J. Black, D.~W. Jacobs, and J.~Malik, ``End-to-end recovery of human shape and pose,'' in \emph{Proceedings of the IEEE conference on computer vision and pattern recognition}, 2018, pp. 7122--7131.

\bibitem{SMPL:2015}
M.~Loper, N.~Mahmood, J.~Romero, G.~Pons-Moll, and M.~J. Black, ``{SMPL}: A skinned multi-person linear model,'' \emph{ACM Transactions on Graphics, (Proc. SIGGRAPH Asia)}, vol.~34, no.~6, pp. 248:1--248:16, Oct. 2015.

\bibitem{sigal2007combined}
L.~Sigal, A.~Balan, and M.~Black, ``Combined discriminative and generative articulated pose and non-rigid shape estimation,'' \emph{Advances in neural information processing systems}, vol.~20, 2007.

\bibitem{zuffi2015stitched}
S.~Zuffi and M.~J. Black, ``The stitched puppet: A graphical model of 3d human shape and pose,'' in \emph{Proceedings of the IEEE Conference on Computer Vision and Pattern Recognition}, 2015, pp. 3537--3546.

\bibitem{7363472}
P.~Beeson and B.~Ames, ``Trac-ik: An open-source library for improved solving of generic inverse kinematics,'' in \emph{2015 IEEE-RAS 15th International Conference on Humanoid Robots (Humanoids)}, 2015, pp. 928--935.

\bibitem{pavllo20193d}
D.~Pavllo, C.~Feichtenhofer, D.~Grangier, and M.~Auli, ``3d human pose estimation in video with temporal convolutions and semi-supervised training,'' in \emph{Proceedings of the IEEE/CVF conference on computer vision and pattern recognition}, 2019, pp. 7753--7762.

\bibitem{ionescu2013human3}
C.~Ionescu, D.~Papava, V.~Olaru, and C.~Sminchisescu, ``Human3. 6m: Large scale datasets and predictive methods for 3d human sensing in natural environments,'' \emph{IEEE transactions on pattern analysis and machine intelligence}, vol.~36, no.~7, pp. 1325--1339, 2013.

\bibitem{sun2021monocular}
Y.~Sun, Q.~Bao, W.~Liu, Y.~Fu, M.~J. Black, and T.~Mei, ``Monocular, one-stage, regression of multiple 3d people,'' in \emph{Proceedings of the IEEE/CVF international conference on computer vision}, 2021, pp. 11\,179--11\,188.

\bibitem{sun2022putting}
Y.~Sun, W.~Liu, Q.~Bao, Y.~Fu, T.~Mei, and M.~J. Black, ``Putting people in their place: Monocular regression of 3d people in depth,'' in \emph{Proceedings of the IEEE/CVF Conference on Computer Vision and Pattern Recognition}, 2022, pp. 13\,243--13\,252.

\bibitem{sun2023trace}
Y.~Sun, Q.~Bao, W.~Liu, T.~Mei, and M.~J. Black, ``Trace: 5d temporal regression of avatars with dynamic cameras in 3d environments,'' in \emph{Proceedings of the IEEE/CVF Conference on Computer Vision and Pattern Recognition}, 2023, pp. 8856--8866.

\bibitem{kanazawa2019learning}
A.~Kanazawa, J.~Y. Zhang, P.~Felsen, and J.~Malik, ``Learning 3d human dynamics from video,'' in \emph{Proceedings of the IEEE/CVF conference on computer vision and pattern recognition}, 2019, pp. 5614--5623.

\bibitem{kocabas2020vibe}
M.~Kocabas, N.~Athanasiou, and M.~J. Black, ``Vibe: Video inference for human body pose and shape estimation,'' in \emph{Proceedings of the IEEE/CVF conference on computer vision and pattern recognition}, 2020, pp. 5253--5263.

\bibitem{choi2021beyond}
H.~Choi, G.~Moon, J.~Y. Chang, and K.~M. Lee, ``Beyond static features for temporally consistent 3d human pose and shape from a video,'' in \emph{Proceedings of the IEEE/CVF conference on computer vision and pattern recognition}, 2021, pp. 1964--1973.

\bibitem{wei2022capturing}
W.-L. Wei, J.-C. Lin, T.-L. Liu, and H.-Y.~M. Liao, ``Capturing humans in motion: Temporal-attentive 3d human pose and shape estimation from monocular video,'' in \emph{Proceedings of the IEEE/CVF Conference on Computer Vision and Pattern Recognition}, 2022, pp. 13\,211--13\,220.

\bibitem{choutas2022accurate}
V.~Choutas, L.~M{\"u}ller, C.-H.~P. Huang, S.~Tang, D.~Tzionas, and M.~J. Black, ``Accurate 3d body shape regression using metric and semantic attributes,'' in \emph{Proceedings of the IEEE/CVF Conference on Computer Vision and Pattern Recognition}, 2022, pp. 2718--2728.

\bibitem{wang2022tepose}
Z.~Wang and S.~Ostadabbas, ``Live stream temporally embedded 3d human body pose and shape estimation,'' in \emph{arXiv preprint: https://arxiv.org/pdf/2207.12537.pdf}, July 2022.

\bibitem{nam2023cyclic}
H.~Nam, D.~S. Jung, Y.~Oh, and K.~M. Lee, ``Cyclic test-time adaptation on monocular video for 3d human mesh reconstruction,'' in \emph{Proceedings of the IEEE/CVF International Conference on Computer Vision}, 2023, pp. 14\,829--14\,839.

\bibitem{moon2020i2l}
G.~Moon and K.~M. Lee, ``I2l-meshnet: Image-to-lixel prediction network for accurate 3d human pose and mesh estimation from a single rgb image,'' in \emph{European Conference on Computer Vision}.\hskip 1em plus 0.5em minus 0.4em\relax Springer, 2020, pp. 752--768.

\bibitem{zeng2022smoothnet}
A.~Zeng, L.~Yang, X.~Ju, J.~Li, J.~Wang, and Q.~Xu, ``Smoothnet: A plug-and-play network for refining human poses in videos,'' in \emph{European Conference on Computer Vision}.\hskip 1em plus 0.5em minus 0.4em\relax Springer, 2022, pp. 625--642.

\bibitem{shen2023global}
X.~Shen, Z.~Yang, X.~Wang, J.~Ma, C.~Zhou, and Y.~Yang, ``Global-to-local modeling for video-based 3d human pose and shape estimation,'' in \emph{Proceedings of the IEEE/CVF Conference on Computer Vision and Pattern Recognition}, 2023, pp. 8887--8896.

\bibitem{yang2024kitro}
F.~Yang, K.~Gu, and A.~Yao, ``Kitro: Refining human mesh by 2d clues and kinematic-tree rotation,'' in \emph{Proceedings of the IEEE/CVF Conference on Computer Vision and Pattern Recognition}, 2024, pp. 1052--1061.

\bibitem{van2008visualizing}
L.~Van~der Maaten and G.~Hinton, ``Visualizing data using t-sne.'' \emph{Journal of machine learning research}, vol.~9, no.~11, 2008.

\bibitem{johnson2011learning}
S.~Johnson and M.~Everingham, ``Learning effective human pose estimation from inaccurate annotation,'' in \emph{CVPR 2011}.\hskip 1em plus 0.5em minus 0.4em\relax IEEE, 2011, pp. 1465--1472.

\bibitem{hu2013massage}
L.~Hu, Y.~Wang, J.~Zhang, J.~Zhang, Y.~Cui, L.~Ma, J.~Jiang, L.~Fang, and B.~Zhang, ``A massage robot based on chinese massage therapy,'' \emph{Industrial Robot: An International Journal}, vol.~40, no.~2, pp. 158--172, 2013.

\bibitem{li2020enhanced}
C.~Li, A.~Fahmy, S.~Li, and J.~Sienz, ``An enhanced robot massage system in smart homes using force sensing and a dynamic movement primitive,'' \emph{Frontiers in Neurorobotics}, vol.~14, p.~30, 2020.

\bibitem{acupointchineses}
S.~A. of~Traditional Chinese~Medicine, ``Gb 12346-2006 acupoint name and location (national standard of the people’s republic of china),'' \emph{China Standards Press}, 2006.

\bibitem{li2015acupoint}
F.~Li, T.~He, Q.~Xu, L.-T. Lin, H.~Li, Y.~Liu, G.-X. Shi, and C.-Z. Liu, ``What is the acupoint? a preliminary review of acupoints,'' \emph{Pain Medicine}, vol.~16, no.~10, pp. 1905--1915, 2015.

\bibitem{9907739}
B.~Cai, P.~Sun, M.~Li, E.~Cheng, Z.~Sun, and B.~Song, ``An acupoint detection approach for robotic upper limb acupuncture therapy,'' in \emph{2022 12th International Conference on CYBER Technology in Automation, Control, and Intelligent Systems (CYBER)}, 2022, pp. 989--992.

\bibitem{9188367}
L.~Sun, S.~Sun, Y.~Fu, and X.~Zhao, ``Acupoint detection based on deep convolutional neural network,'' in \emph{2020 39th Chinese Control Conference (CCC)}, 2020, pp. 7418--7422.

\end{thebibliography}

\vfill
\end{document}